\documentclass[a4paper,fleqn]{cas-sc}

\usepackage[numbers]{natbib}

\def\tsc#1{\csdef{#1}{\textsc{\lowercase{#1}}\xspace}}
\tsc{WGM}
\tsc{QE}
\tsc{EP}
\tsc{PMS}
\tsc{BEC}
\tsc{DE}
\usepackage{float}
\usepackage{amsmath,amsfonts}
\usepackage{algorithm}
\usepackage{algpseudocode}
\usepackage{array}
\usepackage[caption=false,font=normalsize,labelfont=sf,textfont=sf]{subfig}
\usepackage{textcomp}
\usepackage{url}
\usepackage{verbatim}
\usepackage{graphicx}
\usepackage{cite}
\usepackage{soul}
\usepackage{booktabs}
\usepackage{pifont}
\usepackage{multicol}
\usepackage{multirow}
\usepackage{xcolor}
\usepackage{tcolorbox}
\usepackage{listings}
\usepackage{xspace}
\usepackage{pifont}
\newcommand{\cmark}{\ding{51}}  
\newcommand{\xmark}{\ding{55}}  

\lstdefinelanguage{PDDL}{
    morekeywords={define, domain, requirements, types, predicates, action, parameters, precondition, effect, and, not},
    sensitive=true,
    morecomment=[l]{;},
    morestring=[b]"
}
\lstdefinelanguage{json}{
    basicstyle=\footnotesize\ttfamily,
    numbers=left,
    numberstyle=\tiny,
    stepnumber=1,
    numbersep=5pt,
    showstringspaces=false,
    breaklines=true,
    frame=single,
    backgroundcolor=\color{green!5},
    literate=
     *{0}{{{\color{black}0}}}{1}
      {1}{{{\color{black}1}}}{1}
      {2}{{{\color{black}2}}}{1}
      {3}{{{\color{black}3}}}{1}
      {4}{{{\color{black}4}}}{1}
      {5}{{{\color{black}5}}}{1}
      {6}{{{\color{black}6}}}{1}
      {7}{{{\color{black}7}}}{1}
      {8}{{{\color{black}8}}}{1}
      {9}{{{\color{black}9}}}{1},
}
\definecolor{lightgreen}{rgb}{0.88, 0.95, 0.88}
\definecolor{darkgreen}{rgb}{0.0, 0.5, 0.0}

\begin{document}
\let\WriteBookmarks\relax
\def\floatpagepagefraction{1}
\def\textpagefraction{.001}

\shorttitle{AquaChat++ for Aquaculture Net Pen Inspection Using ROV}

\shortauthors{A. Saad et~al.}

\title[mode=title]{AquaChat++: LLM-Assisted Multi-ROV Inspection for Aquaculture Net Pens with Integrated Battery Management and Thruster Fault Tolerance}

\author[1]{Abdelhaleem Saad}
    \address[1]{Khalifa University Center for Autonomous Robotic Systems (KUCARS), Khalifa University, United Arab Emirates.}
    \ead{abdelhaleem.saad@ku.ac.ae}
    \credit{Conceptualization, Methodology, Software, Writing - Original Draft, Formal Analysis, Writing - Reviews and Editing}

  \author[1]{Waseem Akram}
     \credit{Methodology, Software - Original Draft, Investigation }
     \ead{waseem.akram@ku.ac.ae}


\author[1]{Irfan Hussain}
     \credit{Methodology, Writing - Original Draft, Investigation }
     \cortext[cor1]{Corresponding author: I. Hussain (irfan.hussain@ku.ac.ae)}
\begin{abstract}
Inspection of aquaculture net pens is essential for ensuring the structural integrity and sustainable operation of offshore fish farming systems. Traditional methods, typically based on manually operated or single-ROV systems, offer limited adaptability to real-time constraints such as energy consumption, hardware faults, and dynamic underwater conditions. This paper introduces AquaChat++, a novel multi-ROV inspection framework that uses Large Language Models (LLMs) to enable adaptive mission planning, coordinated task execution, and fault-tolerant control in complex aquaculture environments.
The proposed system consists of a two-layered architecture. The high-level plan generation layer employs an LLM, such as ChatGPT-4, to translate natural language user commands into symbolic, multi-agent inspection plans. A task manager dynamically allocates and schedules actions among ROVs based on their real-time status and operational constraints, including thruster faults and battery levels. The low-level control layer ensures accurate trajectory tracking and integrates thruster fault detection and compensation mechanisms.
By incorporating real-time feedback and event-triggered replanning, AquaChat++ enhances system robustness and operational efficiency. Simulated experiments in a physics-based aquaculture environment demonstrate improved inspection coverage, energy-efficient behavior, and resilience to actuator failures. These findings highlight the potential of LLM-driven frameworks to support scalable, intelligent, and autonomous underwater robotic operations within the aquaculture sector.

\end{abstract}


\maketitle

\section{Introduction}


The global demand for aquaculture has surged over the past decade, driving the expansion of offshore fish farming systems such as net pens \citep{subasinghe2009global,FAO2022}. These structures, while effective for large-scale fish production, are continuously exposed to harsh marine environments that can degrade structural integrity, compromise biosecurity, and increase the risk of fish escape or environmental contamination \citep{int2}. As a result, regular and reliable inspection of aquaculture net pens is critical to ensuring operational safety, productivity, and regulatory compliance \citep{int1}. 

Recent advances in underwater robotics, control systems, and computer vision have enabled significant progress in autonomous inspection \citep{akram2021visual,akram2025enhancing}. Remotely Operated Vehicles (ROVs), in particular, offer a practical platform for deploying sensing payloads such as cameras, sonars and performing close-range inspection in confined underwater environments \citep{lee2022autonomous}. However, most existing ROV-based systems operate in isolation, with limited autonomy and minimal adaptability to dynamic conditions such as power constraints, actuator degradation, and evolving mission demands \citep{salin2018aquaculture,int5}. Moreover, mission planning and coordination typically require expert operators, limiting the scalability and responsiveness of these systems in real-world aquaculture operations \citep{paspalakis2020automated,int6,int4}.

\begin{figure}[h]
    \centering
    \includegraphics[width=0.5\linewidth]{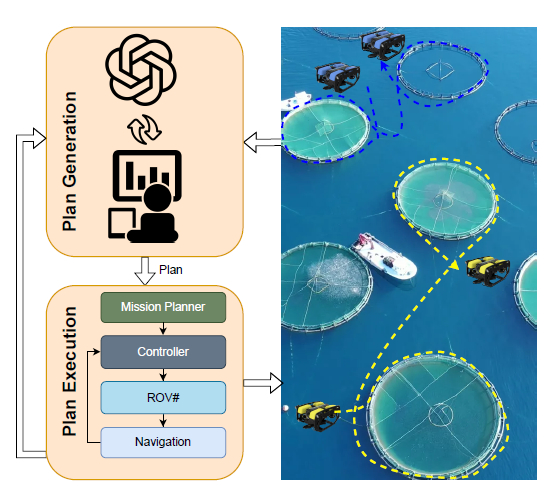}
    \caption{Overview of the AquaChat++ multi-ROV framework: An LLM model interprets human instructions to generate symbolic inspection plans (left), which are executed via a hierarchical control stack comprising a mission planner, controller, and navigation module for multiple ROVs (right).}
    \label{fig:proj}
\end{figure}

To address these challenges, we propose AquaChat++, a novel framework that combines the reasoning capabilities of Large Language Models (LLMs) with multi-ROV coordination, battery-aware mission planning, and fault-tolerant control \citep{akram2023autonomous,llmsmart1}. Unlike traditional inspection pipelines that rely on fixed scripts or manual supervision, AquaChat++ enables natural language-driven task planning and dynamic allocation across multiple ROVs. The proposed system as shown in Figure~\ref{fig:proj}, contains a hierarchical architecture: a high-level LLM-based planner interprets human directives into symbolic plans; a mid-level task manager distributes actions based on battery status, mission priority, and ROV availability; and a low-level control layer ensures robust trajectory tracking with integrated thruster fault detection and compensation \citep{achiam2023gpt}.

The key contributions of this work are as follows:

\begin{itemize}
    \item We introduce a multi-agent inspection framework that leverages LLMs for natural language-driven mission specification and adaptive planning in aquaculture environments.
    
    \item We design a battery-aware task allocation mechanism that maximizes inspection coverage while respecting energy constraints across heterogeneous ROVs.

    \item We integrate a thruster fault-tolerant control strategy that detects and compensates for actuator degradation in real time. During planning, the LLM proactively accounts for thruster faults.

    \item We validate the proposed framework through extensive simulation experiments, demonstrating improved inspection efficiency, energy utilization, and system resilience compared to baseline approaches.
\end{itemize}

The remainder of this paper is organized as follows: Section~\ref{sec:rw} reviews related work in underwater inspection, multi-robot coordination, and LLMs in robotics. Section~\ref{sec:pf} describes the system architecture and key modules of AquaChat++. Section~\ref{sec:sim} details the experimental setup and evaluation metrics. Section~\ref{sec:res} presents results and discussion. Finally, Section~\ref{sec:conc} concludes the paper and outlines directions for future research.

\section{Related Work}\label{sec:rw}

\subsection{Underwater Robotic Inspection in Aquaculture}

Autonomous underwater inspection is pivotal in modern aquaculture, particularly for monitoring the integrity of net pens and ensuring the health and safety of fish farms \citep{wu2025nanodet}. Traditional inspection methods often rely on human divers or ROVs following pre-programmed paths, which can be labor-intensive and lack scalability. Recent research has focused on enhancing the autonomy and efficiency of these systems \citep{yan2025sonar,ge2025navigation}.

For instance, Akram et al. \citep{akram2022visual,akram2023evaluating} developed a deep learning-assisted system for real-time detection of net defects using ROVs, demonstrating high accuracy even under challenging underwater conditions. Similarly, Fabijanić et al. \citep{fabijanic2023autonomous} proposed an autonomous visual inspection framework capable of estimating biofouling buildup on fish pens, reducing the reliance on manual inspections.

Advancements in underwater simultaneous localization and mapping (SLAM) have also contributed to improved autonomy. Pan et al. \citep{pan2025russo} introduced RUSSO, a robust SLAM system that fuses stereo camera, inertial measurement unit (IMU), and imaging sonar data to achieve accurate localization in visually degraded underwater environments. This integration enhances the reliability of autonomous navigation for inspection tasks \citep{cardaillac2025learning,job12leveraging}.

Moreover, the integration of multimodal sensor data has been explored to improve inspection capabilities. Botta et al. \citep{botta2024framework} presented a framework combining vision and acoustic sensors to enhance localization and mapping in complex underwater environments, specifically targeting fish farming applications \citep{ohrem2025solaqua}.

Despite these advancements, challenges remain in achieving fully autonomous, adaptive inspection systems capable of real-time decision-making in dynamic aquaculture environments. The integration of LLM for high-level planning and decision-making, as proposed in our AquaChat++ framework, represents a significant step towards addressing these challenges by enabling intelligent inspection strategies.

\subsection{Large Language Models (LLMs) in Marine Robotics}

LLMs have recently emerged as powerful tools for enabling natural language understanding, task planning, and reasoning across various domains, including robotics \citep{obi2025safeplan}. Their application in marine robotics, however, is still in its early stages. LLMs such as GPT, PaLM, and LLaMA have demonstrated remarkable zero-shot and few-shot generalization capabilities, making them suitable for high-level human-robot interaction and symbolic reasoning in complex, unstructured environments \citep{touvron2023llama,brown2020language,asuzu2025human}.

In terrestrial robotics, LLMs have been used to convert user commands into executable plans, integrate world models for contextual understanding, and perform long-horizon task decomposition \citep{ahn2022can}. Notably, projects like OceanChat \citep{yang2023oceanchat} and Code-as-Policies \citep{liang2023code} have illustrated how LLMs can generate grounded robotic actions from language inputs, forming a natural interface for non-expert users.

Transferring these capabilities to underwater robotics presents unique challenges due to the harsh and dynamic nature of marine environments, limited communication bandwidth, and sensor uncertainty. Early efforts to integrate LLM-based planning into marine robotics are limited but promising. For example, \citep{yang2024oceanplan} proposed a conceptual framework where natural language inputs guide AUV operations, although full-stack implementations remain under development. Similarly, \citep{lee2025wmnav} explored using vision-language models for semantic navigation in partially structured underwater scenes, demonstrating initial success in aligning mission objectives with perceptual inputs. More recent work by \citep{llmsmart2} highlights the potential of generative agents in marine robotics for policy generation and multimodal decision-making. Additionally, \citep{wen2025tinyvla} proposed a lightweight vision-language-action model designed for resource-constrained marine robots, enabling task grounding through compact representations. These studies highlight a growing interest in leveraging LLMs and foundation models for natural language interfacing, adaptive mission planning, and semantic scene understanding in marine robotic applications.

The proposed AquaChat++ framework builds on these ideas by incorporating a layered architecture in which an LLM-driven planner translates high-level language instructions into symbolic inspection plans, coordinating multiple ROVs while considering battery constraints and thruster degradation. This represents a novel fusion of language-based cognition with marine robotic control, enabling intelligent, scalable, and resilient underwater operations within aquaculture sector.

\subsection{Bridging the Gap: AquaChat++ Framework}
Building upon recent advances in LLM-driven planning and marine robotics, we propose AquaChat++, a novel LLM-based multi-agent framework explicitly designed for aquaculture robotics. Unlike existing general-purpose or single-agent solutions, AquaChat++ integrates domain-specific language understanding with adaptive, multi-agent feedback loops. This approach significantly improves robustness and precision in dynamic underwater aquaculture environments by leveraging specialized domain knowledge for accurate contextual interpretation, utilizing intermediate environmental feedback to dynamically adjust actions, and coordinating multiple specialized agents to ensure robust real-time adaptability. AquaChat++ uniquely fills the gap in current literature by providing an integrated solution explicitly tailored to the specialized operational demands of aquaculture net inspections.

\section{Proposed Framework}\label{sec:pf}
This section describes the architecture and key components of the AquaChat++ framework. The system is organized into a hierarchical two-layered architecture designed to enable intelligent multi-ROV inspection missions with energy-aware planning and thruster fault tolerance.

\subsection{System Overview}
The framework is designed as a real-time, LLM-guided inspection system that transforms high-level natural language commands into precise ROV navigation and inspection tasks. It ensures robust execution through a closed-loop feedback mechanism, enabling dynamic adaptation to environmental conditions and task progress.

\begin{figure}
    \centering
    \includegraphics[width=\linewidth]{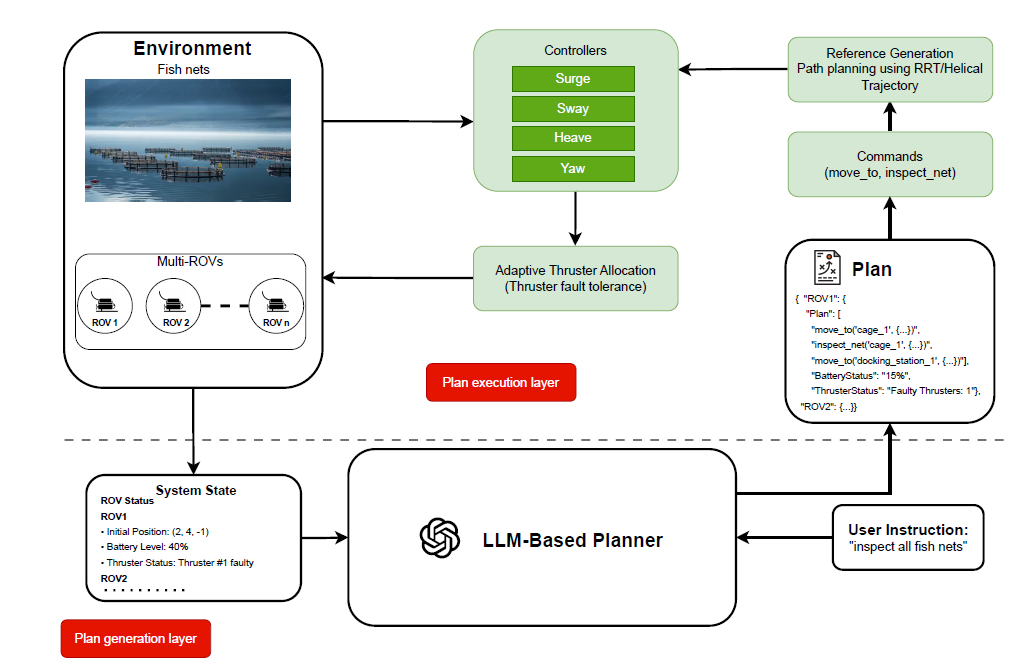}
    \caption{Overview of the LLM-guided inspection framework, comprising a Plan Generation Layer (for symbolic reasoning and validation) and a Plan Execution Layer (for low-level control and fault tolerance).}
    \label{fig:framework_overview}
\end{figure}

The proposed framework is structured into two key layers: the \textit{Plan Generation Layer} and the \textit{Plan Execution Layer}, as illustrated in Figure~\ref{fig:framework_overview}. The Plan Generation Layer takes user instructions and system state information. The user instructions are directions to the ROV for the specific net inspection mission. For example, the user may ask, ``go to cage-1 and inspect the whole net''. On the other side, the system state information includes the environmental details (e.g., net pen geometry, positions of cages and docking stations) and ROV operational constraints (e.g., thruster fault tolerance, battery levels, current 3D position). The LLM-Based Planner takes these inputs and generates a symbolic task plan, and verifies their feasibility under constraints such as battery capacity and thruster functionality.

The second layer named \textit{Plan Execution Layer}, translates the validated symbolic plans into actionable low-level commands e.g. \textit{move\_to, inspect\_net}. For navigation between cages or docking stations, Rapidly-exploring Random Trees (RRT) generate collision-free paths, while helical paths ensure systematic coverage during net inspections. This layer also includes an Adaptive Thruster Allocation module to handle thruster faults and ensure full motion control, and a Controller module responsible for regulating surge, sway, heave, and yaw. This separation between symbolic reasoning and physical control ensures modularity, robustness, and adaptability to real-time environmental changes.

The inspection workflow begins with generating a valid plan for each ROV. The ROVs then navigate to the desired position near the cage via RRT-planned paths. Subsequently, they follow a helical path from top-to-bottom or bottom-to-top, capturing camera data for net inspection. The controller ensures accurate path tracking using state feedback.

In the following subsequent section, we discuss each layer in detail

\begin{figure}[!t]

\begin{tcolorbox}[
    colback=lightgreen,
    colframe=darkgreen,
    arc=0mm,
    boxrule=0.5pt,
    fonttitle=\bfseries,
    width=\textwidth,
    left=2pt,
    right=2pt,
    top=2pt,
    bottom=2pt,
    boxsep=2pt,
    before upper=\scriptsize\linespread{0.8}\selectfont  
]

You are an AI-central planner tasked with coordinating multiple Remotely Operated Vehicles (ROVs) for inspecting multiple cylindrical aquaculture net pens in an underwater environment. The number of ROVs, cages, and docking stations is provided by the environment status.Your objective is to create efficient, and feasible inspection plans for each ROV while strictly follow the constraints.

\medskip

\textbf{Environment Status}

\textbf{Net Cage Information}
   
\begin{itemize}
   \item Shape: Cylindrical, Diameter: 5 meters, Depth range: 0 meters (surface) to -5 meters
\end{itemize}

\textbf{Net Cage Positions}

Cage 1: (0, 0), Cage 2: (20, 20), Cage 3: (20, -20),  
Cage 4: (-20, 20), Cage 5: (-20, -20)

\textbf{Docking Stations}

 Docking Station 1: (10, 0), Docking Station 2: (25, 25)



\begin{minipage}[t]{0.48\textwidth}
\textbf{- ROV1}
\begin{itemize}
    \item Initial Position: (2, 4, -1)
    \item Battery Level: 40\%
    \item Thruster Status: Thruster \#1 faulty
\end{itemize}
\end{minipage}
\hfill
\begin{minipage}[t]{0.48\textwidth}
\textbf{- ROV2}
\begin{itemize}
    \item Initial Position: (25, 10, -2)
    \item Battery Level: 100\%
    \item Thruster Status: All thrusters functional
\end{itemize}
\end{minipage}
<\textbf{Thruster Configuration Matrix}>


\textbf{Constraints \& Rules (STRICTLY FOLLOW)}

\begin{itemize}
    \item \textbf{Inspection Distance}: ROVs must maintain EXACTLY a 3-meter separation from the cage center during inspections to prevent collisions. Only two possible options add 3 in x-axis or subtract 3 in y-axis. You should decide which minimizes distance.
    \item \textbf{Battery}:
    \begin{itemize}
        \item Each cage inspection consumes approximately 25\% battery.
        \item An ROV must dock BEFORE reaching a critical battery level (below 10\%) to ensure sufficient battery for go docking.
        \item If a cage inspection will cause the battery to be below 10\%, ROV must dock instead
    \end{itemize}
    \item \textbf{Thruster}:
    \begin{itemize}
        \item Evaluate thruster faults to determine if the ROV can safely complete inspections.
        \item we consider the motion ONLY in surge, sway, heave and yaw
        \item If deemed insufficient thrust control exists, instruct the ROV to dock immediately for repairs.
        \item If The working thrusters can generate motion in surge, sway, heave and yaw, you can use ROV for inspection.
    \end{itemize}
    \item \textbf{Docking}:
    \begin{itemize}
        \item All ROVs must dock at the nearest docking station upon completion of their assigned tasks. Each ROV should dock at different stations unless number of ROVs more than the number of docking stations
        \item If User task is not completed, ROVs can dock and then complete the task
    \end{itemize}
    \item \textbf{Inspection Direction}:
    \begin{itemize}
        \item ROVs do not have to inspect the same number of cages; assign cages based on proximity to minimize total distance.
        \item Alternate vertical inspection direction to minimize vertical movements, for example: top-to-bottom (0\,m to -5\,m) for the first cage, then bottom-to-top (-5\,m to 0\,m) for the next, and so on.

    \end{itemize}
\end{itemize}

\textbf{Functions Available}
    \begin{verbatim}
move_to(target, {'x': value, 'y': value, 'z': value})
inspect_net(target, {'direction': 'top-to-bottom'/'bottom-to-top', 'method': 'standard', 'distance': 3})
* Distance is measured from ROV position and center of the cage
\end{verbatim}

\textbf{Output Format}

Understand the spatial relation between different objects in the environment. Provide the plan in JSON format and do not include any additional explanation inside or outside the JSON. The format consists of main keys: \texttt{ROV1} and \texttt{ROV2}. Each key must include:
\begin{itemize}
    \item \texttt{Plan}: A sequential list of actions using the available functions above.
    \item \texttt{BatteryStatus}: Estimated remaining battery percentage after plan execution.
    \item \texttt{ThrusterStatus}: Summary of thruster functionality (e.g., "Faulty Thrusters: 1, 2, 4", "All Thrusters Functional").
\end{itemize}

\textbf{User Instruction}

Inspect all fish nets. Generate the inspection plan now.
\end{tcolorbox}
\caption{LLM-Based Planner Prompt Structure.}
\label{fig:prompt}
\end{figure}
\subsection{Plan Generation Layer}

The LLM-Based Planner constitutes the core of the plan generation layer of the proposed frameworks, facilitating the translation of abstract user commands into actions defined in the main system prompt. By using LLM, the planner ensures seamless interpretation, decomposition, and contextual adaptation of tasks for robust aquaculture net pen inspection. This section details the components and functionalities of the LLM-Based Planner.

The main objective of the LLM-Based Planner is to process high-level user instructions, such as ``Inspect all fish nets''. The LLM parses these instructions, integrating them with predefined contextual information about the environment and the status of each ROV. This prompt includes: (i) geometric and spatial information about the aquaculture environment, such as the shape, diameter, and depth of each cylindrical net pen, as well as their precise $(x, y)$ coordinates; (ii) the positions of docking stations; (iii) the initial states of each ROV, including their 3D positions, battery levels, and thruster status; (iv) the full 6$\times$6 thruster configuration matrix, used to determine the ROVs' ability to perform motion in surge, sway, heave, and yaw; (v) system-level constraints and safety rules, such as minimum allowable battery thresholds, inspection offsets, and fault tolerance policies; and (vi) the set of executable functions available to the planner, including \texttt{move\_to()} and \texttt{inspect\_net()}, along with a detailed description of their expected parameters and usage. This rich prompt structure enables the LLM to reason over environment dynamics, task requirements, and robot limitations, and to generate feasible, efficient action sequences. The complete prompt template is illustrated in Figure~\ref{fig:prompt}.


The output is a symbolic plan that decomposes the high-level user instruction into discrete, executable tasks aligned with the mission objectives. For instance, a command such as ``Inspect all fish nets'' is transformed into a set of sub-tasks that may include navigating to designated net pens, performing structured inspections at a fixed offset, and returning to a docking station. These tasks are explicitly encoded in a structured JSON format, where each ROV is assigned a sequential \texttt{Plan} composed of symbolic function calls.

The LLM translates each sub-task into a corresponding symbolic action using predefined control commands such as \texttt{move\_to()} and \texttt{inspect\_net()}. These actions are defined with exact parameters, including spatial coordinates, inspection direction (top-to-bottom or bottom-to-top), and offset distance (e.g., 3 meters from the net pen center). The symbolic plan also includes the estimated \texttt{BatteryStatus} after plan execution, and a summary of the ROV’s \texttt{ThrusterStatus} to reflect any thruster faults considered during planning.

The generated plan adheres strictly to operational constraints, including spatial constraints, minimum energy requirements, and motion feasibility derived from the thruster configuration matrix. For example, the LLM accounts for whether the ROV can maneuver effectively in surge, sway, heave, and yaw before assigning inspection tasks. The planner ensures that the resulting sequence of actions satisfies both the environment and the vehicle's limitations, thus enabling robust execution in realistic underwater conditions.

By encoding these high-level objectives into parameterized symbolic operations, the LLM-Based Planner effectively bridges the gap between abstract user commands and low-level robotic execution. An example of such a symbolic output is illustrated in Figure~\ref{fig:full-plan}, showing action sequences for two ROVs including motion paths, inspection actions, and docking steps, all generated in accordance with system constraints and mission goals.

\begin{figure}[htp]
\centering
\scriptsize  
\lstset{
    language=json,
    basicstyle=\scriptsize
}
\begin{lstlisting}
{
  "ROV1": {
    "Plan": ["move_to('cage_1', {'x': 3.0, 'y': 0.0, 'z': 0.0})",
      "inspect_net('cage_1', {'direction':'top-to-bottom', 'method':'standard', 'distance':3})",
      "move_to('docking_station_1', {'x': 10.0, 'y': 0.0, 'z': 0.0})"],
    "BatteryStatus": "15%",
    "ThrusterStatus": "Faulty Thrusters: 1"},
  "ROV2": {
    "Plan": ["move_to('cage_2', {'x': 17.0, 'y': 20.0, 'z': 0.0})",
      "inspect_net('cage_2', {'direction':'top-to-bottom', 'method':'standard', 'distance':3})",
      "move_to('cage_3', {'x': 17.0, 'y': -20.0, 'z': 0.0})",
      "inspect_net('cage_3', {'direction':'bottom-to-top', 'method':'standard', 'distance':3})",
      "move_to('cage_4', {'x': -17.0, 'y': 20.0, 'z': 0.0})",
      "inspect_net('cage_4', {'direction':'top-to-bottom', 'method':'standard', 'distance':3})",
      "move_to('cage_5', {'x': -17.0, 'y': -20.0, 'z': 0.0})",
      "inspect_net('cage_5', {'direction':'bottom-to-top', 'method':'standard', 'distance':3})",
      "move_to('docking_station_2', {'x': 25.0, 'y': 25.0, 'z': 0.0})"],
    "BatteryStatus": "0%",
    "ThrusterStatus": "All Thrusters Functional"}
}
\end{lstlisting}
\caption{Example symbolic output plan generated by the LLM-Based Planner, showing per-ROV action sequences, energy consumption estimates, and thruster status summaries. }
\label{fig:full-plan}
\end{figure}

One of the key strengths of the LLM-Based Planner is its ability to dynamically adapt the symbolic plan based on real-time feedback and environmental inputs. During the inspection process, the ROV continuously updates the planner with observations from onboard sensors. For instance, if unexpected obstacles or changes in the net pen structure are detected, the LLM reassesses the plan, reconfiguring tasks to accommodate the new context. This dynamic adaptation mechanism ensures robust mission execution, reducing the likelihood of failure due to unforeseen environmental factors.

\begin{figure}[h]
    \centering
    \includegraphics[width=1\linewidth]{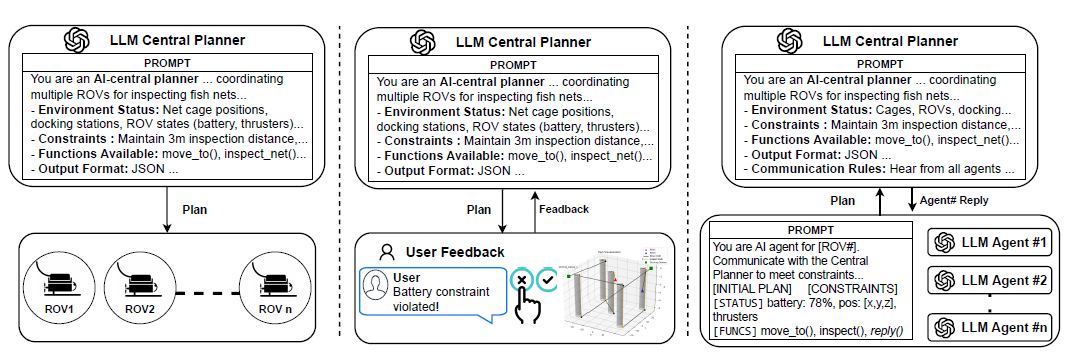}
    \caption{
        Three LLM-based coordination approaches for autonomous aquaculture net pen inspections by multiple ROVs:
        (i) CLP: A single LLM generates plans for all ROVs;
        (ii) HAP: A central LLM adaptively refines plans based on human feedback;
        (iii) MAS: Multiple ROVs coordinate via LLM modules for decentralized planning.
    }
    \label{fig:frameworks}
\end{figure}

To evaluate the generality and flexibility of our LLM-based planning approach, we propose and analyze three distinct approaches (Figure \ref{fig:frameworks}): a centralized single-LLM planner, a human-in-the-loop adaptive planner, and a distributed multi-agent system with local LLM modules per ROV. 

\subsubsection{Central LLM Planner (CLP)}
The baseline architecture utilizes a centralized LLM to interpret high-level user commands, system prompts, and real-time system states, including battery levels and thruster conditions, to formulate comprehensive inspection plans. This architecture translates the complete prompt into actionable ROV commands. Formally, the initial inspection plan is derived as

\begin{equation}
    P_{\text{initial}} = \mathcal{F}_{\text{LLM}}(S_t, I_u, C_b, F_t)
\end{equation}

where $S_t$ represents the current system state, $I_u$ the user instructions, $C_b$ the battery constraints, and $F_t$ the current status of the thrusters. A validation layer subsequently verifies the feasibility of the generated plan to ensure compliance with operational constraints.

\subsubsection{Human-in-the-Loop Adaptive Planner (HAP)}

To address potential shortcomings in purely autonomous planning, we propose an adaptive human-in-the-loop framework. This iterative approach capitalizes on the LLM’s capabilities to refine inspection plans through continuous human feedback. After generating an initial inspection plan, the system dynamically evaluates constraint violations such as battery depletion or thruster faults.

\begin{figure}
    \centering
    \includegraphics[width=0.5\linewidth]{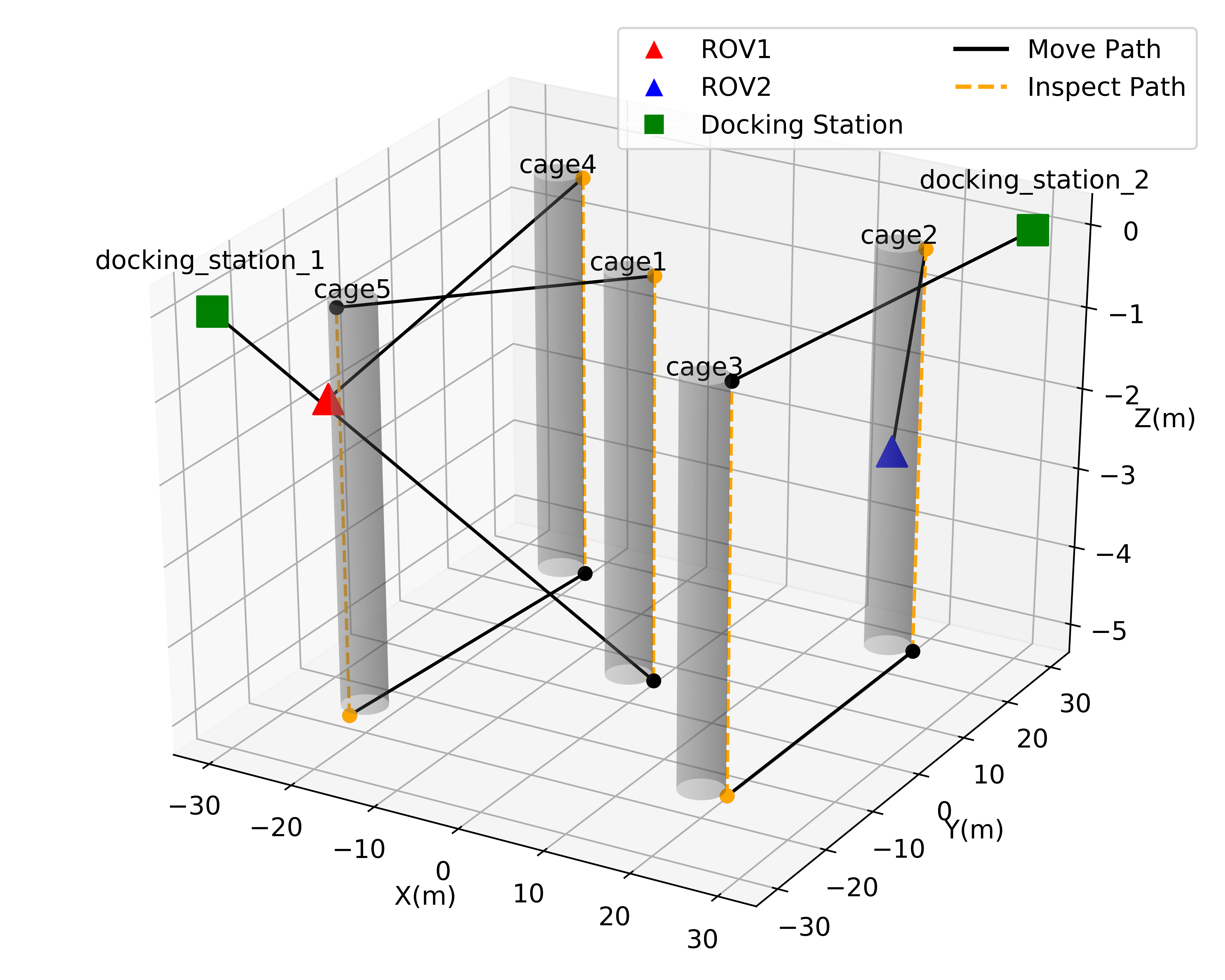}
    \caption{
        3D visualization of the inspection plan showing cages, docking stations, and assigned paths for multiple ROVs. This visualization is shown to users to support decision-making and feedback, based on both spatial interpretation and system validation outputs. 
    }
    \label{fig:PlanVisualization}
\end{figure}

A detailed plan visualization (Figure \ref{fig:PlanVisualization}) is presented to the user, alongside diagnostic outputs from the system’s validation module as shown in Figure~\ref{fig:ValidationOutput}. Based on this visual representation and textual feedback (such as battery levels or actuator failures), users can approve the plan or provide corrective input to the LLM planner:

\begin{equation}
P_{\text{final}} =\mathcal{F}_{\text{LLM}}(P_{\text{initial}}, F_{\text{user}})
\end{equation}

Here, $F_{\text{user}}$ symbolizes human feedback. This process iteratively continues until a viable plan that meets both human approval and system constraints is finalized.

\begin{figure}
\centering
\begin{minipage}{0.9\linewidth}
\footnotesize
\begin{flushleft}
\texttt{Evaluating ROV1...} \\[0.5ex]
\texttt{\textcolor{red}{Error: Battery constraint violated! Battery is now 0.00\%, too low before reaching docking station.}}\\
\texttt{\textcolor{red}{Error: Thruster 1, 2 failures prevent net inspection!}}\\[0.5ex]
\texttt{\textcolor{red}{ROV1 plan is NOT valid! }}\\[1ex]
\texttt{Evaluating ROV2...}\\
\texttt{\textcolor{darkgreen}{ ROV2 plan is VALID.}}\\
\texttt{\textcolor{darkgreen}{Final Battery: 75.00\%}}
\end{flushleft}
\end{minipage}
\caption{Example output from the system validation module. This diagnostic feedback is presented to the user along with the plan visualization to guide iterative corrections.}
\label{fig:ValidationOutput}
\end{figure}

\subsubsection{LLM-Coordinated Multi-Agent System (MAS)}
Extending the capabilities of autonomous and adaptive planning, we propose a sophisticated multi-agent distributed system.
In this framework, each ROV ($ROV_i$) employs a dedicated LLM agent that optimizes local trajectories and compliance with the assigned ROV operational constraints while negotiating with a central planner.

The distributed architecture employs multiple LLM agents to concurrently manage high-level plans and low-level execution details, significantly enhancing adaptability and fault tolerance in complex operational environments. Each agent receives localized system states, processes relevant prompts, and negotiates with the central planner to finalize its execution plan. This centralized communication structure enables scalable and fault-tolerant coordination without requiring direct inter-agent interactions.

The multi-agent system significantly mitigates common pitfalls associated with single-agent architectures, such as hallucinations in long-horizon tasks and rigid adherence to pre-generated plans without considering intermediate environmental feedback. By incorporating environmental feedback after each execution step, our multi-agent system allows ROVs to dynamically adapt inspection paths, efficiently handle intermediate failures, and maintain robust operational integrity throughout the inspection missions.

\subsection{Plan Execution Layer}
The \textit{Plan Execution Layer} is the second layer responsible for the generated symbolic plan. The plan is executed by deploying a multi-ROV concept. Each ROV is given the navigation path. In addition, we also incorporate an adaptive thruster configuration module that is responsible for accommodating the thruster fault scenario and generating the required thrust forces for the navigation task. In the end, we design the control module to allow the ROVs' navigation within the aquaculture environment.

\subsubsection{ROV Model}

In this work, we used BlueROV2 which is a compact and versatile ROV designed for underwater exploration and research. Its dynamic model is derived using the standard 6-DOF equations of motion for marine vessels, incorporating hydrodynamic effects and rigid-body dynamics. The vehicle's parameters, listed in Table~\ref{tab:BlueROV2_parameters}, include mass, moments of inertia, added mass, and damping coefficients.

\begin{table}[htbp]
    \centering
    \caption{BlueROV2 hydrodynamic and rigid-body parameters}
    \label{tab:BlueROV2_parameters}
    \renewcommand{\arraystretch}{1.2}
    \small
    \begin{tabular}{|l|l|r|l|}
        \hline
        \textbf{Parameter} & \textbf{Description} & \textbf{Value} & \textbf{Unit} \\
        \hline
        $m$ & Mass & 11.26 & kg \\
        $I_x$ & Roll inertia & 0.30 & kg$\cdot$m$^2$ \\
        $I_y$ & Pitch inertia & 0.63 & kg$\cdot$m$^2$ \\
        $I_z$ & Yaw inertia & 0.58 & kg$\cdot$m$^2$ \\
        $X_{\dot{u}}$ & Surge added mass & 1.72 & kg \\
        $Y_{\dot{v}}$ & Sway added mass & 0.00 & kg \\
        $Z_{\dot{w}}$ & Heave added mass & 5.47 & kg \\
        $K_{\dot{p}}$ & Roll added mass & 0.00 & kg$\cdot$m$^2$ \\
        $M_{\dot{q}}$ & Pitch added mass & 1.25 & kg$\cdot$m$^2$ \\
        $N_{\dot{r}}$ & Yaw added mass & 0.40 & kg$\cdot$m$^2$ \\
        $X_u$ & Surge linear damping & -11.74 & N$\cdot$s/m \\
        $Y_v$ & Sway linear damping & -20.00 & N$\cdot$s/m \\
        $Z_w$ & Heave linear damping & -31.87 & N$\cdot$s/m \\
        $K_p$ & Roll linear damping & -25.00 & N$\cdot$s/rad \\
        $M_q$ & Pitch linear damping & -44.91 & N$\cdot$s/rad \\
        $N_r$ & Yaw linear damping & -5.00 & N$\cdot$s/rad \\
        $X_{|u|u}$ & Surge quadratic damping & -18.18 & N$\cdot$s$^2$/m$^2$ \\
        $Y_{|v|v}$ & Sway quadratic damping & -21.66 & N$\cdot$s$^2$/m$^2$ \\
        $N_{|r|r}$ & Yaw quadratic damping & -1.55 & N$\cdot$s$^2$/rad$^2$ \\
        \hline
    \end{tabular}
\end{table}

\begin{figure}[htbp]
    \centering
    \includegraphics[width=0.5\linewidth]{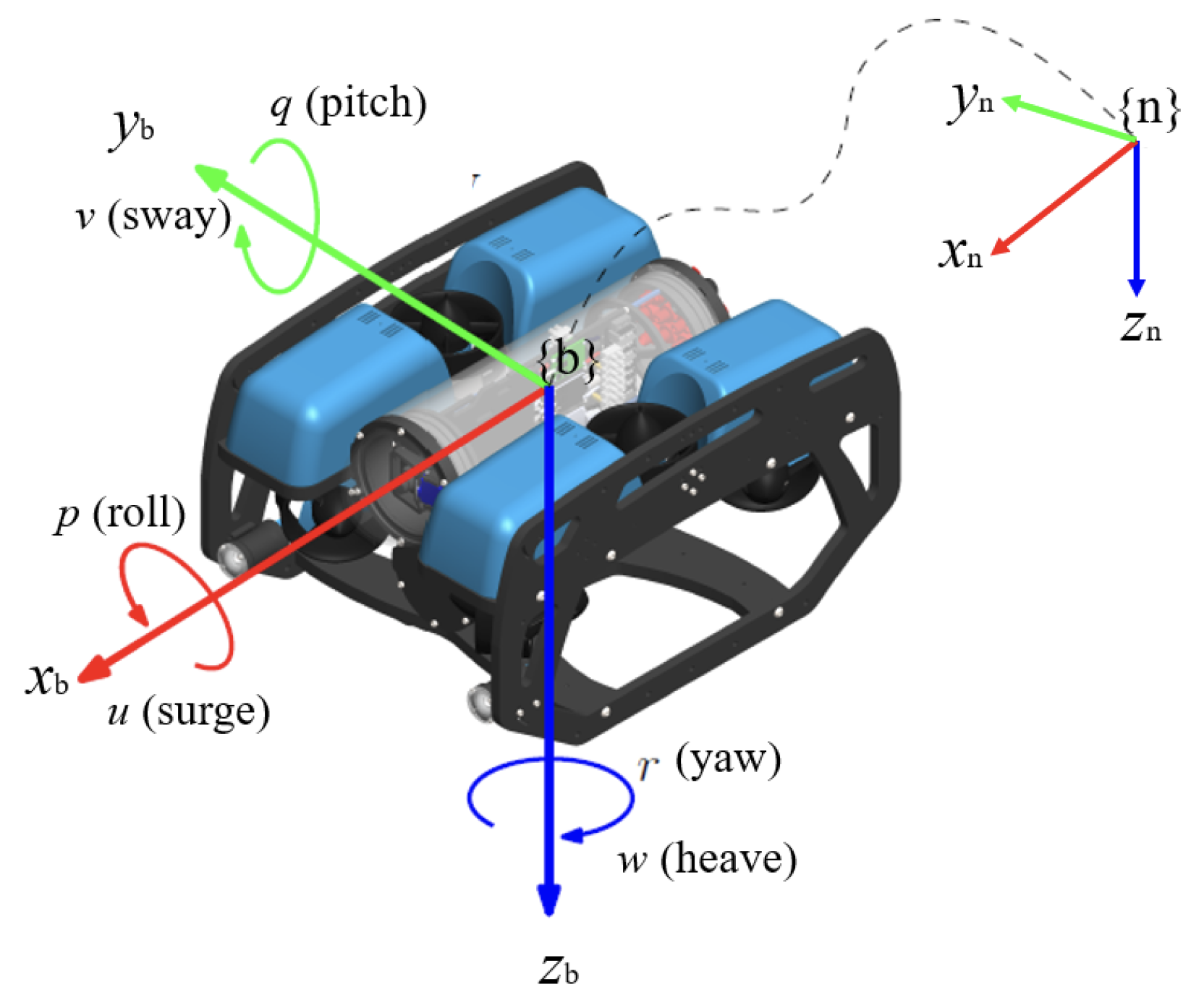}
    \caption{Schematic illustration of the NED, and body-fixed frames used to describe ROV motion. The NED frame is tangent to the Earth’s surface, while the body-fixed frame is aligned with the vehicle’s surge, sway, and heave axes, with roll, pitch, and yaw rotations about each respective axis. (Reproduced from~\citep{ref_frames})}
    \label{fig:ref-frame}
\end{figure}

The motion of the BlueROV2 is described using the North-East-Down (NED) and body-fixed reference frames, as detailed in Figure~\ref{fig:ref-frame}. The NED frame, established as a tangent plane to the Earth's surface, allows for positions to be clearly defined in terms of \(x_n\), \(y_n\), and \(z_n\) coordinates, where \(x_n\) points towards true North, \(y_n\) towards the East, and \(z_n\) directly downwards. The body-fixed reference frame—denoted as \(x_b\), \(y_b\), and \(z_b\)—aligns \(x_b\) in the forward direction, with \(y_b\) towards starboard, \(z_b\) downwards, and its origin at the Center of Gravity (CG). The positive direction of \( \psi \) is defined to rotate clockwise relative to true North.

The generalized pose and velocity vectors are defined as:
\begin{equation}
    \eta = [x, y, z, \phi, \theta, \psi]^T, \quad \nu = [u, v, w, p, q, r]^T,
\end{equation}
where \( \eta \) represents position and orientation, and \( \nu \) denotes linear and angular velocities.

The 6-DoF kinematics equations involve transformations of body-frame velocities to the North-East-Down (NED) world frame using Euler angles. Specifically, the kinematic equation is represented as \( \dot{\eta} = J(\eta)v \), where \( J(\eta) \) is the transformation matrix:

\begin{equation}
\centering
J(\eta) = \begin{bmatrix} R^n_b(\Theta) & 0_{3\times3} \\ 0_{3\times3} & T_\theta(\Theta) \end{bmatrix}
\end{equation}

where \( T_\theta(\Theta) \) converts angular velocities from the body frame to the NED frame, and \( R^n_b(\Theta) \) maps the body-frame linear velocities to the NED frame.

\begin{minipage}[t]{.5\textwidth}
\centering
\( R^n_b(\Theta) = \begin{bmatrix} c_\theta & 0 & -c_\phi s_\theta \\ 0 & 1 & s_\phi \\ s_\theta & 0 & c_\phi c_\theta \end{bmatrix} \)
\end{minipage}
\hfill
\begin{minipage}[t]{.5\textwidth}
\centering
\( T_\theta(\Theta) = \begin{bmatrix} 1 & s_\phi \tan(\theta) & c_\phi \tan(\theta) \\ 0 & c_\phi & -s_\phi \\ 0 & s_\phi / c_\theta & c_\phi / c_\theta \end{bmatrix} \)
\end{minipage}

where \( c \) and \( s \) denote \( \cos() \) and \( \sin() \).

According to Fossen \citep{fossen2011handbook}, the 6-DOF equations of motion are defined as:
\begin{equation}
   M_{RB} \dot{\nu} + M_{A} \dot{\nu} + C_{RB}(\nu) \nu + C_{A} (\nu) \nu + D(\nu)\nu + g(\eta) = \tau_{thrust} + \tau_d
   \label{eq:dynamiceq1}
\end{equation}
Here, \( \nu\) denotes the body-frame velocities; \( M_{RB}\) and \( M_{A}\) represent the rigid-body and added mass matrices, respectively. \( C_{RB}(\nu)\) and \( C_{A}(\nu)\) are the rigid-body and added Coriolis and centripetal matrices, while \( D(\nu) \) is the drag matrix including linear and nonlinear terms, and \( g(\eta) \) represents the generalized restoring forces, such as buoyancy and gravity. \( \tau_{thrust} \) refers to control inputs, and \( \tau_d \) to disturbance forces like wind and surface currents.

The matrices \( M_{RB} \), \( M_A \), \( C_{RB} \), \( C_A \), and \( D \) are constructed using the parameters from Table~\ref{tab:BlueROV2_parameters}, as detailed in Appendix~\ref{app:matrices}.

\subsubsection{Path Planning}\label{sec:rrt}


The RRT* algorithm is employed to generate collision-free 3D paths for ROVs navigating complex underwater environments. It is specifically used to plan safe paths toward target locations such as fish nets (for inspection) or docking stations, while avoiding obstacles including other fish nets that are considered obstacles.

To achieve this, the planner samples the 3D workspace, grows the tree toward the goal, and rewires to minimize cost while respecting obstacle constraints. After a valid path is found, a post-processing step shortens and smooths the path to ensure efficient navigation.

\begin{algorithm}
    \caption{RRT* with Path Shortening for Obstacle-Free Path Generation}
    \label{alg:rrtstar}
    \begin{algorithmic}[1]
        \State Sample \( q_{\text{rand}} \) in obstacle-free 3D space
        \State Identify nearest node \( q_{\text{near}} \) in the existing tree
        \State Extend toward \( q_{\text{rand}} \) with a fixed step size \( \eta \)
        \If{\( q_{\text{new}} \) is collision-free (no fish nets or cage obstacles)}
            \State Add \( q_{\text{new}} \) to the tree
            \State Rewire the tree to minimize the cost-to-come
        \EndIf
        \State Perform path shortening by attempting to connect non-consecutive waypoints with collision-free edges as shown in Figure \ref{fig:rrt}:
        \begin{itemize}
            \item Randomly sample two points along the path (by distance) and compute their positions via interpolation.
            \item Check if the straight edge between the points is collision-free.
            \item If valid, replace the intermediate path segments with the straight edge.
            \item Repeat for a fixed number of iterations.
            \end{itemize}
    \end{algorithmic}
\end{algorithm}

\begin{figure}
    \centering
    \includegraphics[width=0.6\linewidth]{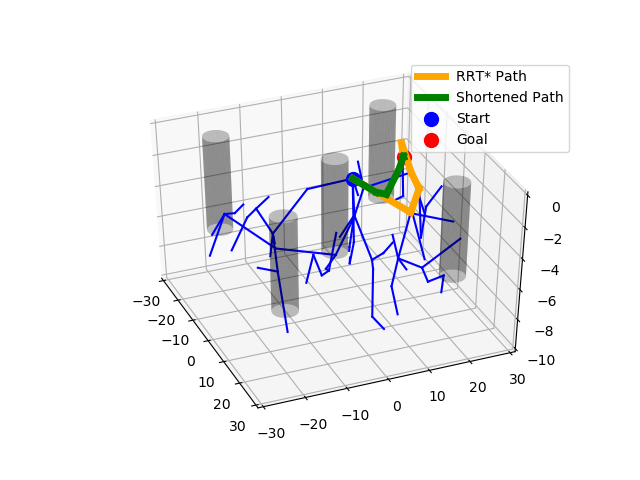}
    \caption{RRT* (green) vs. shortened path (orange) in a 3D environment with fish nets represented as cylindrical obstacles.}
    \label{fig:rrt}
\end{figure}

\vspace{1em}


The helical trajectory ensures full coverage of cylindrical net structures. Given a start point \( (x_0, y_0, z_0) \), radius \( R \), and pitch \( \Delta z \), the parametric equations are defined as:

\begin{equation}
    \begin{cases}
        x(t) = x_c + R \cos(\omega t + \theta_0), \\
        y(t) = y_c + R \sin(\omega t + \theta_0), \\
        z(t) = z_0 + \dfrac{\Delta z}{2\pi} \omega t,
    \end{cases}
    \label{eq:helix}
\end{equation}

where \( \omega = \dfrac{2\pi n_{\text{turns}}}{T} \). Parameters are tuned to match the net’s dimensions and the sensor's field of view (FoV), e.g., \( R = \text{net radius} + 0.5\,\text{m} \) for safety clearance.


\subsubsection{Adaptive Thrust Allocation}

Adaptive thruster allocation module operates at the low-level of the ROV’s control architecture, ensuring precise execution of navigation and inspection tasks despite potential thruster faults. This module integrates adaptive thrust allocation with real-time control to dynamically adjust thrust allocation across available thrusters. Its primary objective is to enable the ROV to accurately follow the reference paths generated by high-level planners in response to commands such as \texttt{move\_to} or \texttt{inspect\_net}. By continuously compensating for thruster inefficiencies or failures, the adaptive control system maintains stable and reliable maneuvering, ensuring that mission objectives are achieved even under degraded conditions.


The BlueROV2 is equipped with six thrusters arranged in a standard configuration to provide 4-DOF control (surge, sway, heave, and yaw). The thrust allocation matrix \( K \) maps individual thruster commands \( u = [T_1, T_2, \dots, T_6]^T \) to the desired generalized forces \( \tau = [X, Y, Z, K, M, N]^T \):

\begin{equation}
    \tau = K u.
\end{equation}

\begin{figure}
    \centering
    \includegraphics[width=0.2\linewidth]{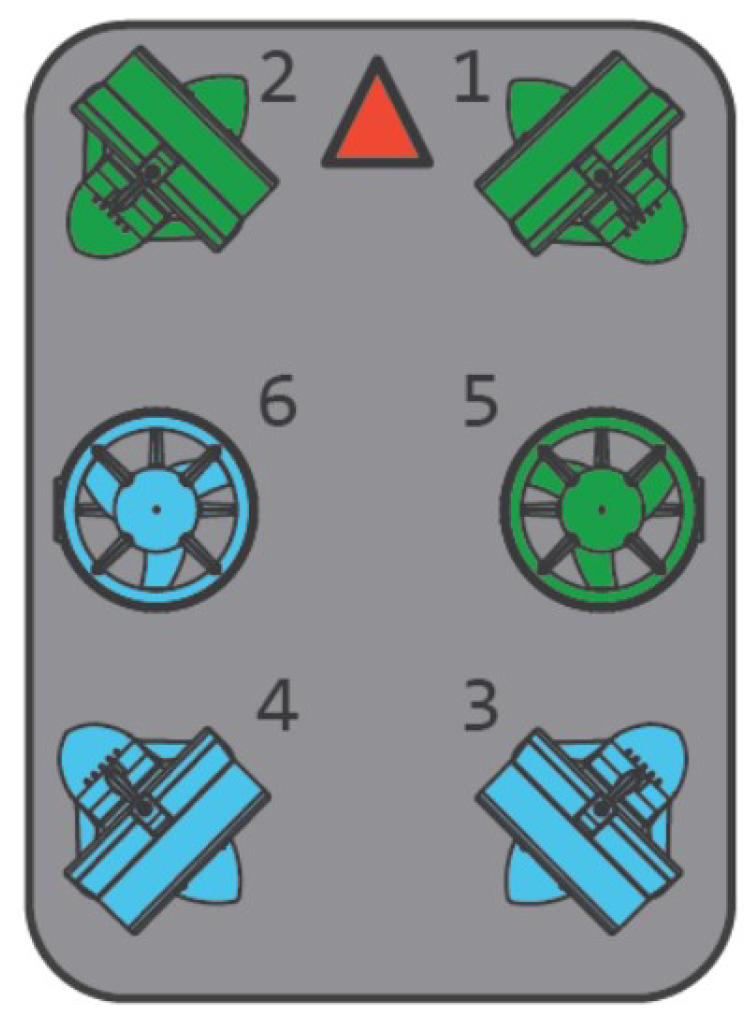}
    \caption{BlueROV2 thruster configuration showing arrangement and orientation of the six thrusters \citep{jmse10121898}.}
    \label{fig:thruster_configuration}
\end{figure}

The thruster arrangement is shown in Figure~\ref{fig:thruster_configuration}, with positions defined relative to the vehicle's center of gravity (CG). Each thruster is modeled with first-order dynamics to account for actuator response delays:

\begin{equation}
    \tau_i \dot{T}_i(t) + T_i(t) = T_{i,\text{desired}}(t),
\end{equation}

where \( \tau_i \) is the time constant for the \( i \)-th thruster.

The thrust allocation matrix \( K \) is derived from the geometric configuration of the thrusters. For thruster 1 (and similarly for others), the contribution to body-frame forces and moments is:

\begin{equation}
\begin{bmatrix}
\cos\alpha \\ -\sin\alpha \\ 0 \\
\sin\alpha \cdot l_{z1} \\ \cos\alpha \cdot l_{z1} \\ -\sin\alpha \cdot l_{x1} - \cos\alpha \cdot l_{y1}
\end{bmatrix} t_1
\end{equation}

where \(\alpha \) is the thruster angle and \(l_{x1}, l_{y1}, l_{z1}\) are moment arms. The complete thrust allocation matrix for the BlueROV2 is:

\begin{equation}
K =
\begin{bmatrix}
0.707 & 0.707 & -0.707 & -0.707 & 0 & 0 \\
-0.707 & 0.707 & -0.707 & 0.707 & 0 & 0 \\
0 & 0 & 0 & 0 & 1 & 1 \\
0.051 & -0.051 & 0.051 & -0.051 & 0.111 & -0.111 \\
0.051 & 0.051 & -0.051 & -0.051 & 0.002 & -0.002 \\
-0.167 & 0.167 & 0.175 & -0.175 & 0 & 0
\end{bmatrix}
\end{equation}

The thrust allocation problem requires solving an optimization problem that efficiently distributes forces while accounting for thruster constraints and potential faults. While the pseudo-inverse method provides a basic solution:

\begin{equation}
    u = T^\dagger \tau,
\end{equation}

a more robust approach incorporates thruster degradation and fault tolerance through constrained optimization. The allocation problem is formulated as:

\begin{equation}
\begin{aligned}
& \underset{u}{\text{minimize}}
& & \|Tu - \tau\|_2 \\
& \text{subject to}
& & u_{min} \cdot d \preceq u \preceq u_{max} \cdot d
\end{aligned}
\end{equation}

where:
\begin{itemize}
\item $u \in \mathbb{R}^6$ is the thruster force vector
\item $\tau \in \mathbb{R}^6$ is the desired generalized force
\item $T \in \mathbb{R}^{6\times6}$ is the thrust allocation matrix
\item $d \in [0,1]^6$ represents thruster degradation factors
\item $u_{min}, u_{max}$ are minimum and maximum thruster limits
\end{itemize}

The optimization problem is solved in real-time using convex optimization techniques. For thrusters experiencing complete failure ($d_i = 0$), the corresponding elements are constrained to zero. The implementation handles partial degradation ($0 < d_i < 1$) by scaling the effective thrust limits.

\begin{algorithm}
\caption{Adaptive Thruster Allocation}
\begin{algorithmic}[1]
\State Input: $\tau_{desired}$, $K$, $u_{max}$, $u_{min}$, $d$
\State Define $u \in \mathbb{R}^6$ \Comment{Optimization variable}
\State $objective \gets \text{minimize}(\|Ku - \tau_{desired}\|_2)$
\State $constraints \gets [u_{min} \cdot d \leq u \leq u_{max} \cdot d]$
\State Solve optimization problem with $objective$ and $constraints$
\State Return $u_{optimal}$
\end{algorithmic}
\end{algorithm}

This adaptive allocation approach provides a simple yet effective solution that automatically handles thruster faults and degradation while respecting physical constraints. The method works by continuously adjusting thruster commands based on each thruster's health status - completely eliminating failed thrusters from the allocation scheme and scaling outputs for degraded thrusters according to their remaining capacity. The optimization redistributes forces to functioning thrusters while prioritizing the most critical degrees of freedom needed for vehicle control. By formulating the problem as a constrained minimization problem, the solution maintains real-time performance while ensuring all control inputs stay within each thruster's current operational limits, providing robust fault tolerance without complex heuristic rules. The system automatically compensates for lost thrust capacity by optimally utilizing remaining healthy thrusters, focusing first on maintaining essential motions like depth control and heading stability when facing partial thruster failures.

\subsubsection{Controller Design}
\label{sec:controller}
The BlueROV2's 4-DOF motion (surge, sway, heave, yaw) is controlled via a cascaded PID architecture, chosen for its simplicity and robustness in underwater environments \citep{fossen2011handbook}. 

The PID control law for each DOF is:
\begin{equation}
    u(t) = K_p e(t) + K_i \int_0^t e(\tau) \,d\tau + K_d \frac{de(t)}{dt},
    \label{eq:pid}
\end{equation}
where \( u(t) \) denotes the control input, \( e(t) = r(t) - y(t) \) represents the tracking error, and \( K_p \), \( K_i \), and \( K_d \) are the proportional, integral, and derivative gains, respectively. These gains are individually tuned for each degree of freedom to achieve the desired response.

The tracking error vector \( \mathbf{e}(t) \) encapsulates the difference between the desired and current states across all controlled DOFs, and is expressed as:

\begin{equation}
\mathbf{e}(t) = 
\begin{bmatrix}
e_x(t) \\
e_y(t) \\
e_z(t) \\
e_{\theta}(t)
\end{bmatrix}
=
\begin{bmatrix}
x_d - x(t) \\
y_d - y(t) \\
z_d - z(t) \\
\theta_d - \theta(t)
\end{bmatrix},
\end{equation}

where \( x_d, y_d, z_d, \theta_d \) denote the desired positions and yaw angle, and \( x(t), y(t), z(t), \theta(t) \) represent the corresponding actual states at time \( t \).

Using Lyapunov theory, the closed-loop error dynamics for surge (similarly for other DOFs) are:
\begin{equation}
    \dot{V}(e) = -e^T Q e \leq 0, \quad Q = Q^T > 0,
    \label{eq:lyapunov}
\end{equation}
guaranteeing asymptotic stability under bounded disturbances.

\section{Simulation Setup}\label{sec:sim}

In this section, we describe the simulation environment and its development. To evaluate the performance of the AquaChat++ framework for multi-ROV inspection of aquaculture net pens, we developed a realistic underwater simulation environment using the ROS-Gazebo framework. This environment integrates three key components: the UUV Simulator~\citep{Manhaes_2016}, the BlueROV2 Simulator~\citep{bluerov2_sim, bluerov2_simulator}, and a custom aquaculture net pen model~\citep{akram2022visual}. This setup provides a realistic testbed for validating the perception, navigation, planning, and control algorithms required in aquaculture inspection missions.

The UUV Simulator~\citep{Manhaes_2016} is a Gazebo-based package designed specifically for underwater robotics research. It offers accurate hydrodynamic modeling that captures the effects of drag, buoyancy, and added mass—critical for replicating real underwater dynamics. It also supports various marine sensors including sonars, cameras, DVLs, and depth sensors, enabling end-to-end testing of robotic capabilities in visually and physically complex environments.

The BlueROV2 was chosen as the underwater inspection vehicle. For this purpose, we employed the BlueROV2 Simulator~\citep{bluerov2_simulator}, which is built on top of the UUV Simulator. It provides a detailed vehicle model with accurate hydrodynamic and rigid-body specifications. Additionally, it supports the ability to enable or disable individual thrusters, making it well-suited for evaluating fault-tolerant control strategies under thruster failure conditions.

The aquaculture net pen model was based on the structure proposed in~\citep{akram2022visual}. The model includes a spherical mesh net supported by cylindrical poles and vertical beams to replicate the geometric complexity of real aquaculture installations. The mesh structure and cage layout offer realistic challenges in terms of visual occlusion, trajectory planning, and inspection coverage.

\begin{figure}
    \centering
    \includegraphics[width=\textwidth]{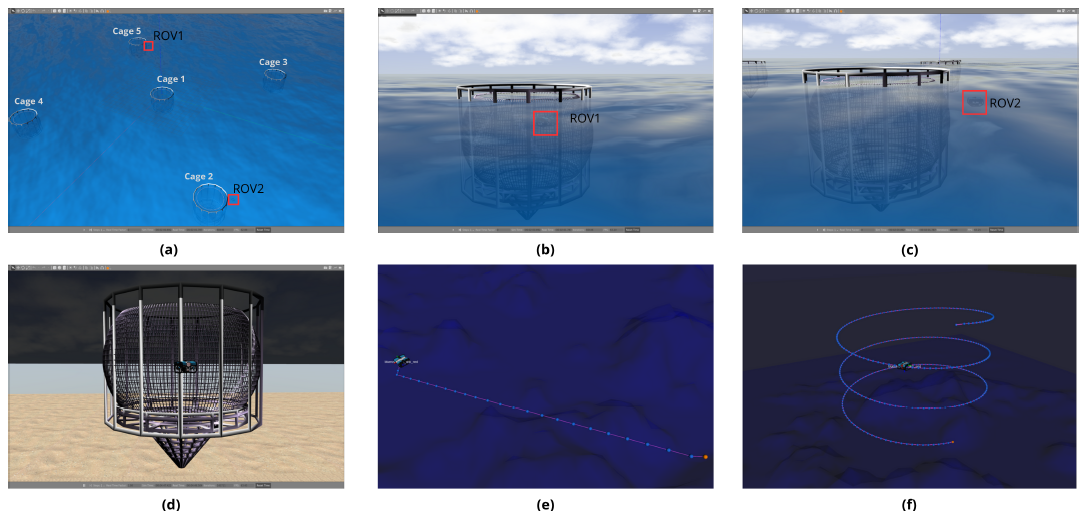}
    \caption{Simulation setup and inspection analysis in Gazebo. (a) Full simulation view at a fixed timestep, with ROV1 and ROV2 inspecting separate cages (detailed in b and c, respectively). (d) Close-up of the net pen structure. (e) Example of move path that avoids obstacles (net pens) , visualized using rviz. (f) Desired inspection trajectory in rviz.}
    \label{fig:sim_setup}
\end{figure}

As illustrated in Figure~\ref{fig:sim_setup}, the simulation environment comprises five net cages and two ROVs operating in parallel. Subfigures (a)–(c) present snapshots captured at a single time instance, demonstrating concurrent inspection operations where ROV1 and ROV2 are each inspecting different net cages. The net pen model used in the simulation was designed in Blender and integrated into the Gazebo simulator in a URDF-compatible format. Subfigure (d) shows the physical structure of the net pen, while subfigures (e) and (f) visualize the executed ROV trajectories, representing move path and inspection path, respectively.

The Gazebo environment allows multiple ROVs to be spawned simultaneously, and multiple net pens are included in the world model. By leveraging ROS, we enable seamless communication with each ROV and independently control their actions, supporting scalable, parallel multi-agent execution.

\section{Results and Discussions}\label{sec:res}

In this section, we present the experimental results of the AquaChat framework. First, we assess the performance of the LLM planner with various custom prompts and evaluate its responses. Then, we demonstrate the ROV's performance in executing the actions outlined in the generated plan during the inspection process.

\subsection{Prompt-Based Evaluation}

To evaluate the planning capabilities of the proposed LLM-based , we simulate multi-ROV net cage inspection in an aquaculture environment consisting of five cages symmetrically distributed in a grid layout. The positions of the cages are as follows: Cage 1 at $(0, 0)$, Cage 2 at $(20, 20)$, Cage 3 at $(20, -20)$, Cage 4 at $(-20, 20)$, and Cage 5 at $(-20, -20)$. Each cage is assumed to be 5 meters in diameter and vertically spans from the surface to a depth of \(-5\) meters. There are two ROVs (ROV1 and ROV2) and two docking stations at two different locations, as shown in Figure~\ref{fig:sim_setup}(a). Each ROV is assigned inspection tasks across different cages under realistic constraints such as battery capacity and thruster degradation.

We used GPT-4o~\citep{openai2024gpt4o} to power each LLM agent in the three planning approaches. Table~\ref{tab:llm-prompts} presents a prompt-based evaluation of CLP, HAP, and MAS. The prompts test each system's ability to generate valid, constraint-compliant, and executable multi-ROV plans. We conducted 20 runs for each prompt using all three planning approaches and reported their planning success rates, execution success rates, and average plan generation time. Importantly, our results confirm that the execution success rate reaches $100\%$ for all approaches—as long as the generated plan is valid.

The CLP planner operates in a single-shot manner, issuing one query per scenario. It performs well in straightforward tasks, such as those in Cases 2 and 5 (Table~\ref{tab:llm-prompts}), where fixed task assignments or linear inspection flows are sufficient. In these cases, CLP is the fastest approach due to its one-shot nature. However, it exhibits significant limitations in more complex scenarios involving stringent constraints or fault conditions—such as battery limitations or thruster failures—resulting in complete failure in Cases 3 and 4.

\begin{table}[t]
\centering
\caption{Comparison of LLM Planning in Multi-ROV Net Cage Inspection Tasks. PSR: planning success rate, EXESR: execution success rate, Time: plan generation time (s)}
\label{tab:llm-prompts}
\resizebox{\textwidth}{!}{
\begin{tabular}{|p{0.5cm}|p{5.3cm}|p{0.8cm} p{0.8cm} p{0.8cm}|p{0.8cm} p{0.8cm} p{0.8cm}|p{0.8cm} p{0.8cm} p{0.8cm}|}
\hline
\textbf{Case} & \textbf{Prompt Description} 
& \multicolumn{3}{c|}{\textbf{CLP}} 
& \multicolumn{3}{c|}{\textbf{HAP}} 
& \multicolumn{3}{c|}{\textbf{MAS}} \\
& & PSR \% & EXESR \% & Time (s) 
& PSR \% & EXESR \% & Time (s) 
& PSR \% & EXESR \% & Time (s) \\
\hline
1 & Inspect all five cages using ROV1 and ROV2 with intermediate docking if needed 
& 60 & 60 & 3.1
& 100 & 100 & 98.0
& 100 & 100 & 13.5 \\
\hline
2 & Assign ROV1 to inspect Cage1, Cage3 and ROV2 to inspect Cage2, Cage4, Cage5 sequentially 
& 90 & 90 & 3.2
& 100 & 100 & 63.5
& 100 & 100 & 9.0 \\
\hline
3 & Plan efficient inspection under 50\% battery constraint for each ROV 
& 0 & 0 & 3.9
& 95 & 95 & 105.0
& 90 & 90 & 18.0 \\
\hline
4 & If ROV2 has a failed vertical thruster, skip deep cages and assign accordingly 
& 0 & 0 & 3.2
& 85 & 85 & 101.5
& 80 & 80 & 18.5 \\
\hline
5 & ROV1 to inspect cages 1--3 and dock, ROV2 to inspect remaining 
& 85 & 85 & 3.5 
& 100 & 100 & 50.3
& 100 & 100 & 12.5 \\
\hline
\end{tabular}
}
\end{table}

HAP planner is the slowest method overall, primarily due to the time taken for human feedback during iterative planning. Despite this, HAP generally requires fewer LLM prompts to converge on a valid plan and achieves the highest overall planning success rate. This makes it the most robust approach when reliability is critical, especially in constraint-heavy or failure cases.

MAS planner typically takes 2 to 4 planning rounds depending on the case complexity. Although slightly less accurate than HAP in some edge cases, MAS achieves comparable performance across most prompts. What sets it apart is that it remains fully autonomous, requiring no human intervention while still handling complex coordination and constraints effectively. This autonomy, combined with good performance and moderate planning time, makes MAS a strong candidate for scalable, real-time multi-ROV deployment.

\subsection{Scenario-Based Evaluation}

In this evaluation, we examine the ability of the three planning  to generate valid and constraint-compliant inspection plans under different battery availability and thruster failure conditions. The following four scenarios are tested:

\begin{itemize}
    \item Scenario 1 (S1) – Battery-Constrained: Number of net inspection exceeds the total battery capacity of all ROVs (each cage requires $25\%$ battery). Intermediate docking is required.
    \item Scenario 2 (S2) – Battery-Sufficient: Number of net inspection is less than the available battery capacity. Docking is only required at the end.
    \item Scenario 3 (S3) – Partial Thruster Failure: The ROV has partial thruster failure but provides enough control for inspection. The valid plan must allow inspection and include docking afterward for repair.
    \item Scenario 4 (S4) – Critical Thruster Failure: The ROV’s remaining thrusters are insufficient to provide full control. The valid plan must skip inspections entirely and perform immediate docking for maintenance.
\end{itemize}

As shown in Table~\ref{tab:scenarios}, all planners successfully produced valid plans under the battery-sufficient condition (S2). However, in the battery-constrained scenario (S1), only HAP and MAS generated valid, constraint-compliant plans. CLP failed to include necessary docking actions, resulting in the violation of the battery constraint.

To evaluate how the  handle partial thruster failures, we define two distinct thruster fault scenarios. In both cases, ROVs experience a loss of one or more thrusters. The system checks whether the remaining thrusters provide sufficient motion control in surge, sway, heave, and yaw. In Scenario 3, all planners correctly assessed the operational capabilities of the ROV and scheduled inspection tasks followed by docking. In Scenario 4, however, only the HAP planner and the MAS planner generated valid plans that instructed immediate docking with no inspections. The CLP planner attempted to assign inspections despite insufficient thruster functionality, violating system constraints.

\begin{table}[H]
\centering
\caption{Plan validity summary for CLP, HAP, and MAS under battery and thruster fault scenarios.}
\label{tab:scenarios}
\begin{tabular}{|c|c|c|}
\hline
\textbf{Scenario} & \textbf{Framework} & \textbf{Valid Plan Generated} \\
\hline
S1 & CLP & \xmark \\
S1 & HAP & \cmark \\
S1 & MAS & \cmark \\
\hline
S2 & CLP & \cmark \\
S2 & HAP & \cmark \\
S2 & MAS & \cmark \\
\hline
S3 & CLP & \cmark \\
S3 & HAP & \cmark \\
S3 & MAS & \cmark \\
\hline
S4 & CLP & \xmark \\
S4 & HAP & \cmark \\
S4 & MAS & \cmark \\
\hline
\end{tabular}
\end{table}

\subsection{Performance of Control Method}

To assess the effectiveness of adaptive thrust allocation in compensating for thruster failures, we conducted two key experiments. The first validates the LLM-generated multi-ROV inspection plan under healthy and faulty thruster conditions.The second focused on the most critical operational mode of our system: the net inspection trajectory. In the context of aquaculture, this trajectory is essential, as it controls the ROV to maintain a consistent and safe standoff distance from the net while navigating vertically from top to bottom in a helical trajectory while heading towards the net.

\subsubsection{Evaluation of Final Inspection Plan}

\begin{figure}[htbp]
    \centering
    \includegraphics[width=0.6\linewidth]{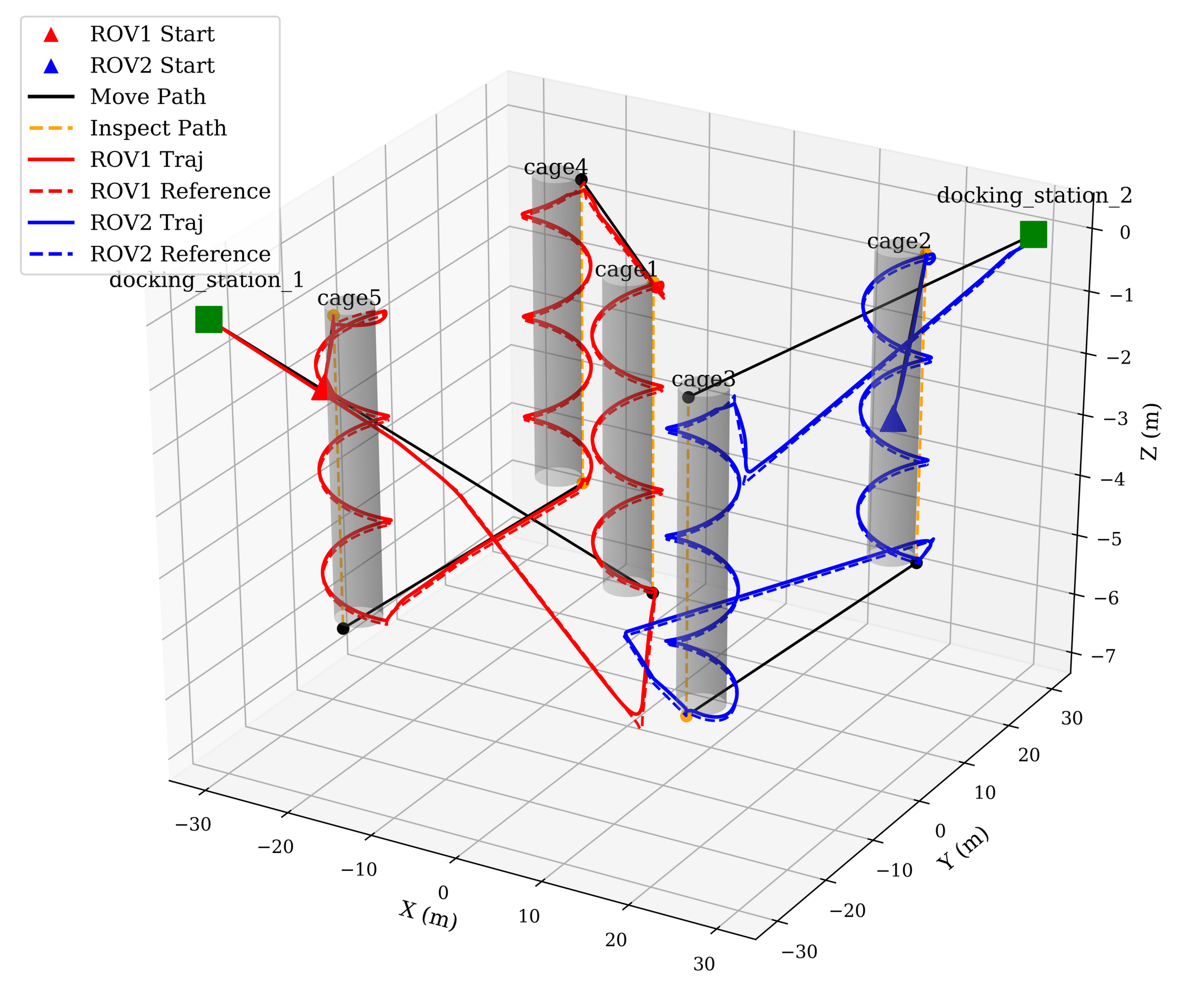}
    \caption{MAS-generated plan executed by ROV1 and ROV2. Red and blue curves represent executed trajectories. ROV1 operates with all thrusters functional; ROV2 executes the inspection tasks with Thrusters 2 and 5 disabled.}
    \label{fig:plan_3d}
\end{figure}

Figure~\ref{fig:plan_3d} illustrates the 3D trajectories of two ROVs executing a complete inspection plan. The MAS planner was selected to generate the inspection plan.

\begin{itemize}
    \item \textbf{ROV1} (red): All thrusters are functional. It follows the sequence: inspect \texttt{cage\_5}, \texttt{cage\_4}, and \texttt{cage\_1}, then proceeds to \texttt{docking\_station\_1}.
    \item \textbf{ROV2} (blue): Thrusters 2 and 5 are disabled. It inspects \texttt{cage\_2} and \texttt{cage\_3}, then returns to \texttt{docking\_station\_2}.
\end{itemize}

\begin{figure}[htbp]
    \centering
    \begin{minipage}[t]{0.48\textwidth}
        \centering
        \includegraphics[width=\linewidth]{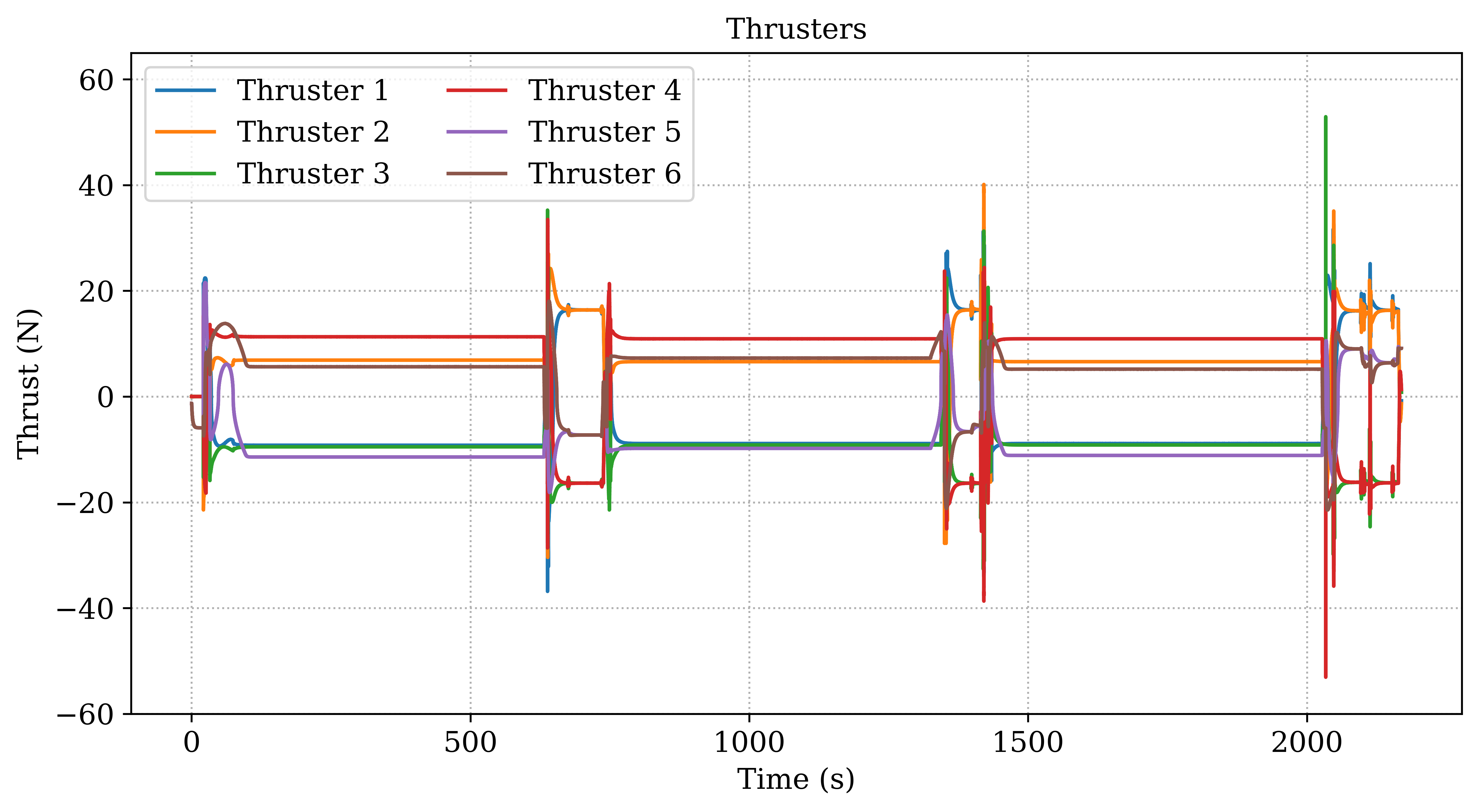}
        \label{fig:thrust_rov1}
    \end{minipage}
    \hfill
    \begin{minipage}[t]{0.48\textwidth}
        \centering
        \includegraphics[width=\linewidth]{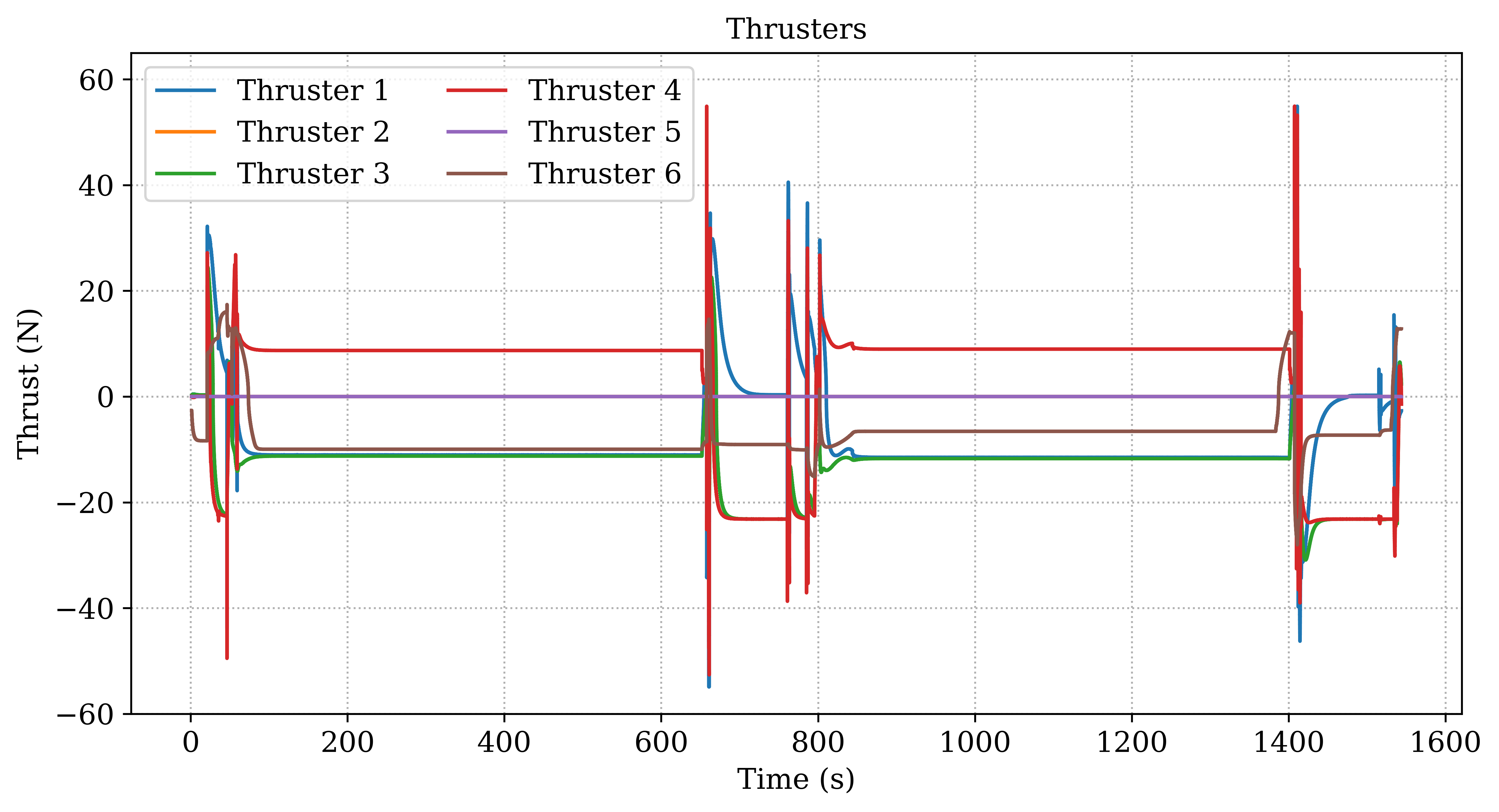}
        \label{fig:thrust_rov2}
    \end{minipage}
    \caption{Thruster profiles of ROV1 (left) with all thrusters functional, and ROV2 (right) operating under partial failure of Thrusters 2 and 5. The controller compensates by redistributing thrust.}
    \label{fig:thrust_comparison}
\end{figure}

\begin{figure}
    \centering
    \begin{minipage}[t]{0.48\textwidth}
        \centering
        \includegraphics[width=\linewidth]{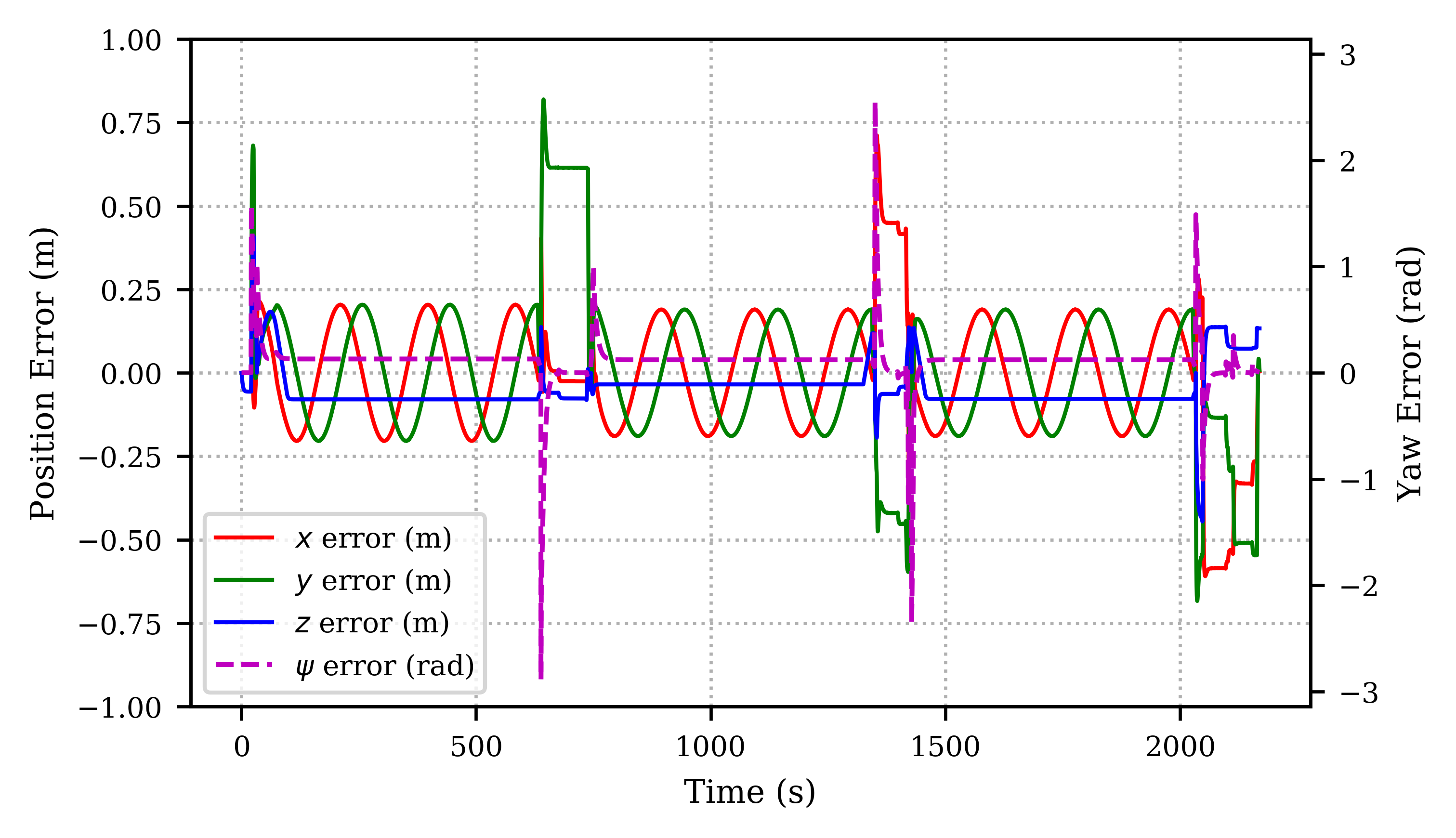}
        \label{fig:error_rov1}
    \end{minipage}
    \hfill
    \begin{minipage}[t]{0.48\textwidth}
        \centering
        \includegraphics[width=\linewidth]{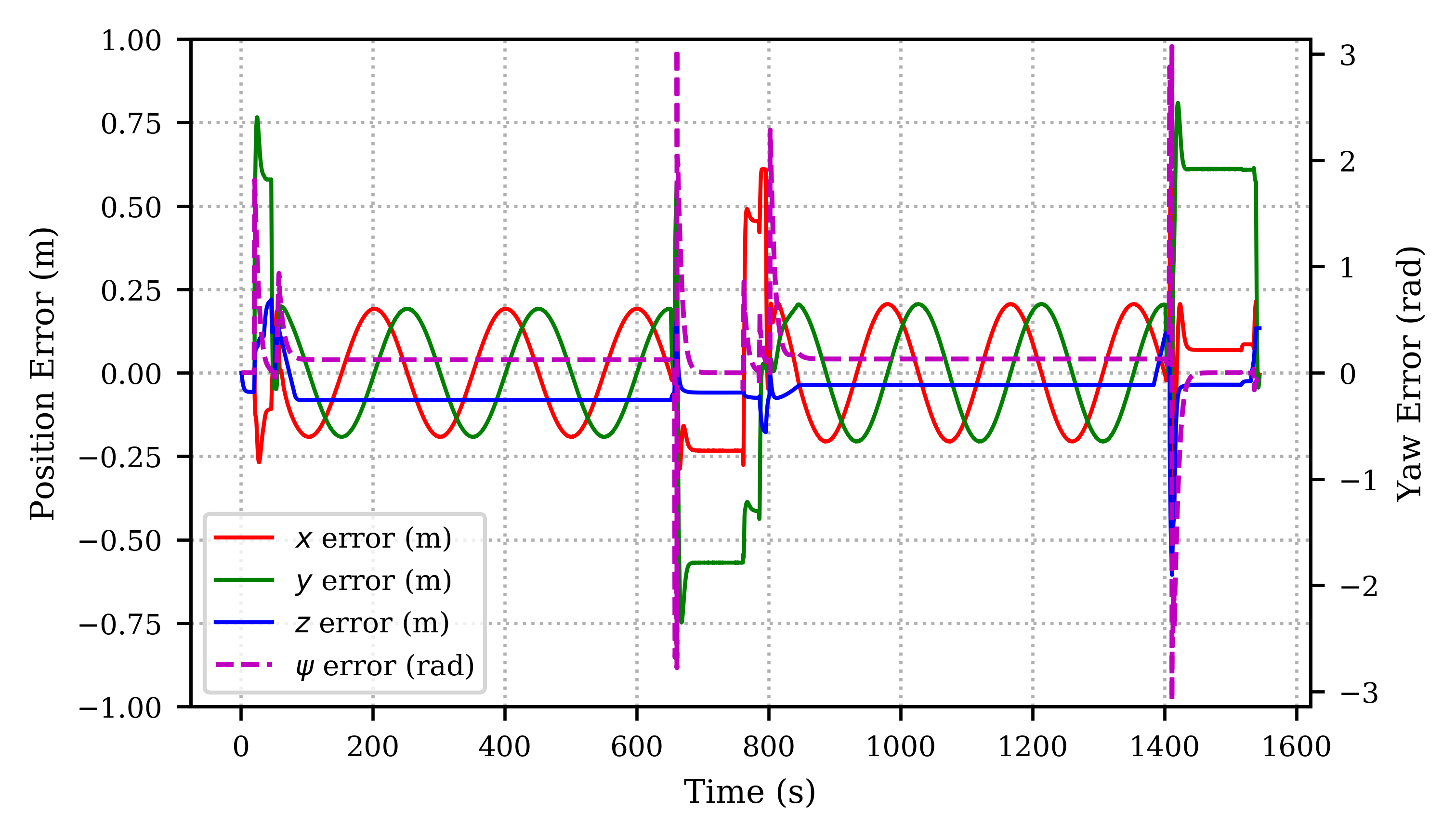}
        \label{fig:error_rov2}
    \end{minipage}
    \caption{Tracking errors in $x$, $y$, $z$, and yaw for ROV1 (left) and ROV2 (right). ROV1 maintains stable control with all thrusters operational, while ROV2, despite faults, shows only slightly increased yaw fluctuations.}
    \label{fig:error_comparison}
\end{figure}

The MAS planner allocated ROV2 to do inspection tasks as the remaining functional thrusters can produce thrust in the 4-DOF (surge, sway, heave, yaw). This demonstrates the planner’s ability to generate a plan under system constraints (thruster faults and battery usage). In contrast, ROV2 operated under partial actuator failure, with Thrusters 2 and 5 disabled. The control system adapted by redistributing the required thrust among the remaining functional thrusters. As shown in Figure~\ref{fig:thrust_comparison}, Thruster 2 and Thruster 5 remain at zero throughout the mission, confirming that the controller did not allocate any thrust for the faulty thrusters. The other thrusters show compensatory increases to produce the desired thrust in all controllable DOFs. For example, Thruster 5 is responsible for heave motion, so its failure requires Thruster 6 to increase its thrust output to maintain depth control. Figures~\ref{fig:error_comparison} present the tracking errors in position and yaw for ROV1 and ROV2, respectively. ROV1 shows minimal tracking errors across all axes, indicating precise trajectory following due to fully functional thrusters. In contrast, ROV2 exhibits slightly higher errors, particularly in yaw, as the control system compensates for the actuator faults while maintaining acceptable performance.

This comparative analysis highlights the effectiveness of the MAS planner in generating fault-aware plans and the controller’s ability to dynamically adapt in case of the presence of thruster faults.



\vspace{1em}  

\subsubsection{Evaluation of Net Inspection under Different Fault Conditions}

The experiment was conducted to assess the ability of the controller—enhanced with adaptive thrust allocation—to maintain accurate trajectory tracking in the presence of thruster faults. Notably, the thruster failures were deliberately selected based on their thrust contribution rather than being random. According to the previously described thruster configuration of the BlueROV2, the selected thrusters are: thrusters 2 and 4, which control surge, sway, and yaw motions, and thruster 5, which controls heave (vertical) motion.

By disabling combinations of these thrusters, we simulate realistic fault conditions that may occur due to actuator failure, cable disconnection, or motor degradation. Four scenarios were evaluated: (i) a healthy condition with all six thrusters functional (denoted as Normal), (ii) a single fault where thruster 2 is disabled (F2), (iii) a dual fault where thrusters 2 and 5 are disabled (F2,5), and (iv) a severe triple fault where thrusters 2, 4, and 5 are disabled (F2,4,5).

\begin{figure}
    \centering
    \includegraphics[width=0.5\linewidth]{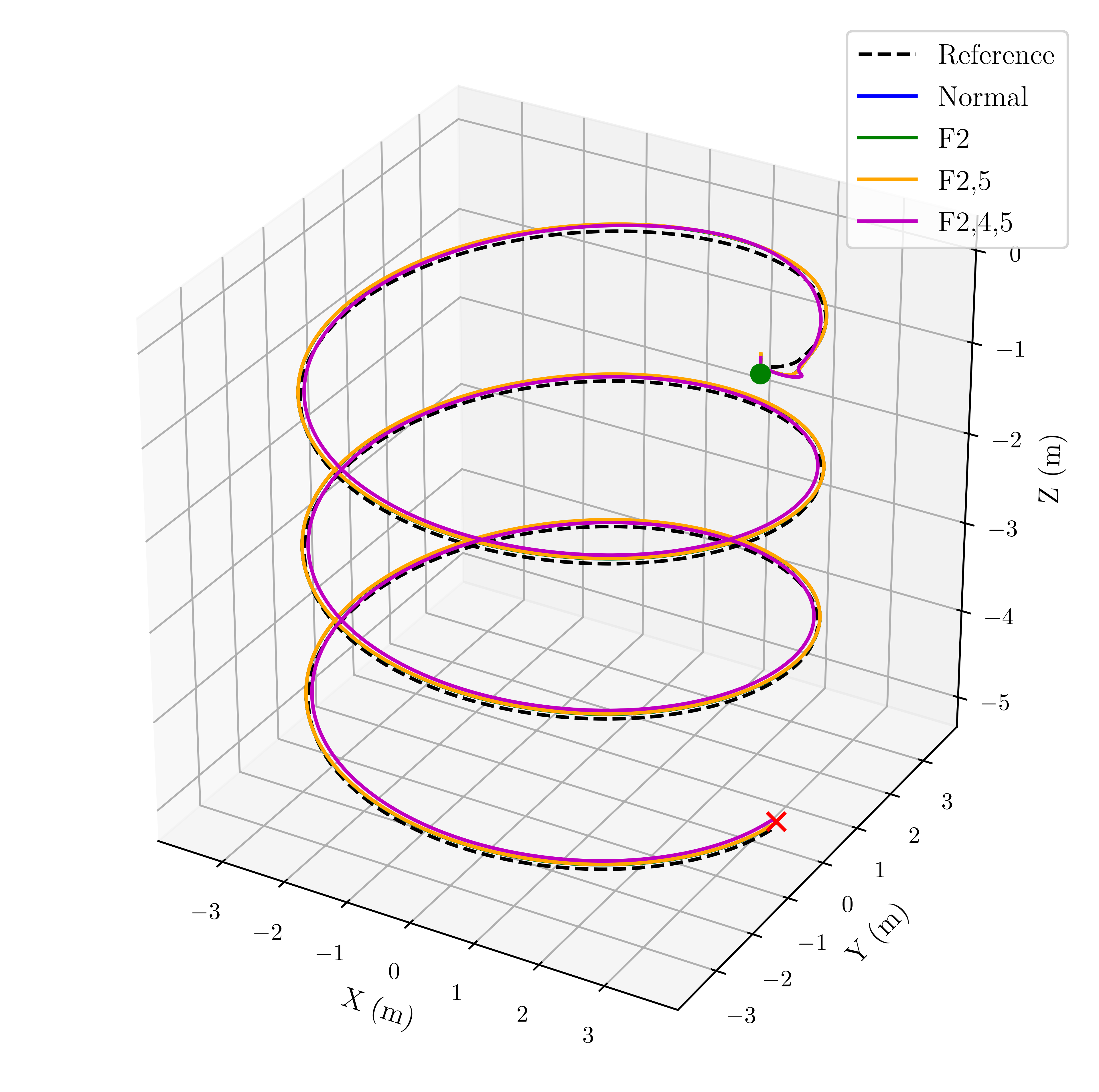}
    \caption{ROV trajectory tracking under different thruster fault conditions using adaptive thrust allocation. The black dashed line represents the reference inspection trajectory. solid lines show ROV trajectory under healthy and faulty scenarios. The green marker indicates the start, and the red cross marks the end.}
    \label{fig:trajectory_comparison}
\end{figure}

Figure~\ref{fig:trajectory_comparison} visualizes the ROV's actual trajectory compared to the reference trajectory under these fault scenarios. All conditions follow a helical descent from $z = 0$~m to $z = -5$~m, mimicking the vertical scanning motion used for detailed net inspection. Colored trajectories represent the ROV's actual path under each condition, while the black dashed line denotes the reference. A green circle marks the initial position, and a red cross denotes the final position.

In the healthy and single-fault cases, the vehicle accurately tracks the path with minor deviations, indicating effective compensation by the control system. Even under the double-fault condition (F2,5), the ROV maintains acceptable tracking in position, although vertical error increases slightly because one of the two heave thrusters is disabled . 

However, in the most severe scenario (F2,4,5), although the ROV maintains position accuracy, it exhibits a significant yaw deviation. This manifests as rotational drift, which severely impacts the visual quality of inspection data. As the onboard camera is front-facing, large yaw errors cause the ROV to deviate from the net, reducing the quality and coverage of the captured imagery—an outcome detrimental to our application.

Table~\ref{tab:average_errors} quantifies the tracking performance across the four scenarios. Positional errors in $x$, $y$, and $z$ remain below $12$~cm, even in degraded conditions. However, yaw error escalates from $\sim0.15$~rad to over $1$~rad in the severe fault case.

\begin{table}
    \centering
    \caption{Average Tracking Errors by Fault Condition}
    \label{tab:average_errors}
    \begin{tabular}{lcccc}
        \toprule
        \textbf{Condition} & \textbf{X (m)} & \textbf{Y (m)} & \textbf{Z (m)} & \textbf{Yaw (rad)} \\
        \midrule
        Normal          & 0.118 & 0.118 & 0.077 & 0.155 \\
        F2              & 0.118 & 0.118 & 0.078 & 0.156 \\
        F2,5            & 0.120 & 0.119 & 0.081 & 0.156 \\
        F2,4,5          & 0.110 & 0.110 & 0.081 & 1.011 \\
        \bottomrule
    \end{tabular}
\end{table}

In summary, the experiment highlights the effectiveness of the adaptive thrust allocation strategy in compensating for partial actuator failures during inspection tasks. The system maintains accurate positional tracking even under severe conditions, such as the failure of three thrusters, allowing the ROV to navigate to a docking station for necessary maintenance. However, the yaw control performance significantly deteriorates in such scenarios, which undermines the feasibility of reliable net inspection. Since precise yaw regulation is critical for maintaining consistent visual coverage of the net surface, the degraded performance limits the operation of the system under extreme thruster faults.

\section{Conclusion}\label{sec:conc}

This work proposed a novel framework named AquaChat, using LLM-guided navigation coupled ROVs for the inspection of aquaculture net pens. By integrating high-level planning with domain-specific tasks and real-time sensing, the system demonstrates the potential for enhancing underwater inspection efficiency and precision. The modular design of the framework enables adaptability to various operational scenarios, addressing challenges such as environmental complexity. Moreover, it offers a user-friendly and ease-of-use aquaculture net pen inspection via LLM planner by taking instruction in a natural language format. Overall, the proposed approach lays a strong foundation for advancing the role of AI-driven technologies in underwater inspection and beyond.
In future, the framework will be enhanced to address challenges including model adaptation and scalability. Addressing these challenges will unlock new opportunities for improving aquaculture management and sustainability. 


\section*{Acknowledgement}
\noindent This work is supported by the Khalifa University under Award No. RC1-2018-KUCARS-8474000136, CIRA-2021-085, MBZIRC-8434000194, KU-BIT-Joint-Lab-8434000534 and KU-Stanford :8474000605.

\section*{Decleration}
During the preparation of this work the author(s) used ChatGPT-4 in order to improve language and readability. After using this tool/service, the author(s) reviewed and edited the content as needed and take(s) full responsibility for the content of the publication.

\printcredits

\bibliographystyle{model1-num-names}

\bibliography{cas-refs}

\appendix
\section{Matrices}
\label{app:matrices}
In this section, we detail the matrices used to model the hydrodynamics of a marine vessel, based on the assumption that movements between the Degrees of Freedom (DoFs) are decoupled. The total mass matrix \(M\) is defined as:

\[
M = M_{RB} + M_A
\]

\[
M_{RB} = \begin{bmatrix}
m & 0 & 0 & 0 & 0 & m z_g \\
0 & m & 0 & 0 & -m z_g & 0 \\
0 & 0 & m & 0 & 0 & 0 \\
0 & 0 & 0 & I_x & 0 & 0 \\
0 & -m z_g & 0 & 0 & I_y & 0 \\
m z_g & 0 & 0 & 0 & 0 & I_z
\end{bmatrix}
\]

\[
M_A = -\text{diag}([X_{\dot{u}}, Y_{\dot{v}}, Z_{\dot{w}}, K_{\dot{p}}, M_{\dot{q}}, N_{\dot{r}}])
\]
The Coriolis and centripetal matrix \(C\) is derived from the mass matrix \(M\) and the generalized velocity vector \(\nu\) as follows:

\[
C = C_{RB}(v) + C_A(v)
\]

\[
C_{RB}(v) = \begin{bmatrix}
0 & 0 & 0 & 0 & m w & -m v \\
0 & 0 & 0 & -m w & 0 & m u \\
0 & 0 & 0 & m v & -m u & 0 \\
0 & m w & -m v & 0 & I_z r & -I_y q \\
-m w & 0 & m u & -I_z r & 0 & I_x p \\
m v & -m u & 0 & I_y q & -I_x p & 0
\end{bmatrix}
\]

\[
C_A(v) = \begin{bmatrix}
0 & 0 & 0 & 0 & -Z_{\dot{w}} w & Y_{\dot{v}} v \\
0 & 0 & 0 & Z_{\dot{w}} w & 0 & -X_{\dot{u}} u \\
0 & 0 & 0 & -Y_{\dot{v}} v & X_{\dot{u}} u & 0 \\
0 & -Z_{\dot{w}} w & Y_{\dot{v}} v & 0 & -N_{\dot{r}} r & M_{\dot{q}} q \\
Z_{\dot{w}} w & 0 & -X_{\dot{u}} u & N_{\dot{r}} r & 0 & -K_{\dot{p}} p \\
-Y_{\dot{v}} v & X_{\dot{u}} u & 0 & -M_{\dot{q}} q & K_{\dot{p}} p & 0
\end{bmatrix}
\]

The damping matrix \(D(v)\) accounts for both linear and quadratic damping:
\[
D(v) = -\text{diag}([X_u + X_{|u|} |u|, Y_v + Y_{|v|} |v|, Z_w + Z_{|w|} |w|, K_p + K_{|p|} |p|, M_q + M_{|q|} |q|, N_r + N_{|r|} |r|])
\]

Finally, the restoring force vector \(g(\eta)\) includes forces due to buoyancy and gravity:
\[ g(\eta) = \begin{bmatrix}
(W - B) \sin \theta \\
-(W - B) \cos \theta \sin \phi \\
-(W - B) \cos \theta \cos \phi \\
z_g W \cos \theta \sin \phi \\
z_g W \sin \theta \\
z_g W \cos \theta \sin \phi
\end{bmatrix} \]

When the center of gravity (CG) and the center of origin (CO) of a marine vessel are not aligned, 
 the transformation between them is achieved through the equation \( M_{CO} = H(r_{bg})^T M_{CG} H(r_{bg}) \). Here, \( H \) is the transformation matrix based on the vector \( r_{bg} \) from the CO to the CG, and is defined as:
\[
H = \begin{bmatrix}
I_{3x3} & S(r_{bg})^T \\
0 & I_{3x3}
\end{bmatrix}
\]
with \( S(r_{bg}) \) being the skew-symmetric matrix of \( r_{bg } \).

Thruster dynamics are a fundamental block in simulating marine vehicles, where thrusters exhibit a delayed response due to actuator limitations and fluid dynamics. This behavior is realistically modeled as a first-order system characterized by a time constant \( \tau \), capturing the gradual convergence of actual thrust \( T(t) \) to the commanded input \( T_{\text{desired}}(t) \). The system is governed by the differential equation:

\begin{equation}
\tau \frac{dT(t)}{dt} + T(t) = T_{\text{desired}}(t)
\end{equation}

\end{document}